\pdfoutput=1
\documentclass[letterpaper]{article} % DO NOT CHANGE THIS
\usepackage{aaai23}  % DO NOT CHANGE THIS
\usepackage{times}  % DO NOT CHANGE THIS
\usepackage{helvet}  % DO NOT CHANGE THIS
\usepackage{courier}  % DO NOT CHANGE THIS
\usepackage[hyphens]{url}  % DO NOT CHANGE THIS
\usepackage{graphicx} % DO NOT CHANGE THIS
\usepackage{natbib}  % DO NOT CHANGE THIS AND DO NOT ADD ANY OPTIONS TO IT
\usepackage{caption} % DO NOT CHANGE THIS AND DO NOT ADD ANY OPTIONS TO IT
\usepackage{algorithm}
\usepackage{algorithmic}

\usepackage{adjustbox}
\usepackage{amsmath}
\usepackage{amsthm}
\usepackage{amsfonts}
\usepackage{cleveref}
\usepackage{lipsum}
\newtheorem{thm}{Theorem}
\crefname{thm}{Theorem}{Theorems}

\crefname{lem}{Lemma}{Lemmas}
\usepackage{mathtools}
\usepackage{multirow}
\usepackage{nicefrac}       % compact symbols for 1/2, etc.
\usepackage{comment}
\usepackage{subfig}

\usepackage{newfloat}
\usepackage{listings}
\DeclareCaptionStyle{ruled}{labelfont=normalfont,labelsep=colon,strut=off} % DO NOT CHANGE THIS
\floatstyle{ruled}
\newfloat{listing}{tb}{lst}{}
\floatname{listing}{Listing}
\nocopyright
\title{Neural Diffeomorphic Non-uniform B-spline Flows} %Neural Diffeomorphic Non-Uniform B-Spline Flows}
\author{
    Seongmin Hong\textsuperscript{\rm 1},
    Se Young Chun\textsuperscript{\rm 1,2}\thanks{Corresponding author}
}
\affiliations{
    \textsuperscript{\rm 1} Department of Electrical and Computer Engineering, Seoul National University, Republic of Korea\\
    \textsuperscript{\rm 2} INMC, Interdisciplinary Program in AI, Seoul National University, Republic of Korea\\
    \{smhongok, sychun\}@snu.ac.kr
}

\begin{document}

\maketitle

\begin{abstract}
Normalizing flows have been successfully modeling a complex probability distribution as an invertible transformation of a simple base distribution. However, there are often applications that require more than invertibility. For instance, the computation of energies and forces in physics requires the second derivatives of the transformation to be well-defined and continuous. Smooth normalizing flows employ infinitely differentiable transformation, but with the price of slow non-analytic inverse transforms. In this work, we propose diffeomorphic non-uniform B-spline flows that are at least twice continuously differentiable while bi-Lipschitz continuous, enabling efficient parametrization while retaining analytic inverse transforms based on a sufficient condition for diffeomorphism. Firstly, we investigate the sufficient condition for $\mathcal{C}^{k-2}$-diffeomorphic non-uniform $k$th-order B-spline transformations. Then, we derive an analytic inverse transformation of the non-uniform cubic B-spline transformation for neural diffeomorphic non-uniform B-spline flows. Lastly, we performed experiments on solving the force matching problem in Boltzmann generators, demonstrating that our $\mathcal{C}^2$-diffeomorphic non-uniform B-spline flows yielded solutions better than previous spline flows and faster than smooth normalizing flows. Our source code is publicly available at \url{https://github.com/smhongok/Non-uniform-B-spline-Flow}.
\end{abstract}

\section{Introduction}

Normalizing flows~\cite{rezende2015variational, papamakarios2021normalizing} model complex probability distributions. Normalizing flows are not only performing the probability density estimation but also sampling from the learned probability distribution. Let $\mathcal{X}$ be a dataset from the true target probability distribution $p_{\mathbf{x}}^\star$. Constructing normalizing flows fits a flow-based model $p_\mathbf{x}$ to the true target distribution $p_{\mathbf{x}}^\star$ using a simple base probability distribution $p_{\mathbf{u}}$ and a diffeomorphic (\textit{i.e.}, invertible and differentiable) mapping $T:\Omega \rightarrow \Omega$ where $\Omega$ is a compact subset of $\mathbb{R}^D$ with the following density transformation: % as follows: 
\begin{equation}\label{eq1.1}
    p_{\mathbf{x}}(\mathbf{x}) = p_{\mathbf{u}}(T^{-1}(\mathbf{x})) \left\lvert \mathrm{det} \frac{\partial T^{-1}}{\partial \mathbf{x}} (\mathbf{x}) \right\rvert.
\end{equation} %requires A normalizing flow to fit , using $p_\mathbf{u}$ and $T$ as follows:
For $\mathcal{X}=\{\mathbf{x}^{(n)}\}_{n=1}^{N}$ where %, and 
$\mathbf{x}^{(1)}, \dots, \mathbf{x}^{(N)}$ are \textit{i.i.d.} samples from $p_{\mathbf{x}}^\star$, %. Then,
normalizing flows are %usually 
trained on $\mathcal{X}$ by minimizing the following negative log-likelihood (NLL):
\begin{equation}
    \mathcal{L}_{\text{NLL}} = -\frac{1}{N}\sum_{n=1}^{N} \log p_\mathbf{x} (\mathbf{x}^{(n)})
    \approx -\mathbb{E}_{\mathbf{x} \sim p_{\mathbf{x}}^\star}[\log p_{\mathbf{x}}(\mathbf{x})]
\end{equation}
where the above approximation becomes true as $N \rightarrow \infty$. Density estimation corresponds to obtaining %$p_{\mathbf{x}}(\mathbf{x})$ or 
$T$ from $\mathbf{x}$ and sampling is referred to getting $\mathbf{x}=T(\mathbf{u})$ from $\mathbf{u} \sim p_{\mathbf{u}}(\mathbf{u})$.

Normalizing flows are often used to model physical systems. For example, one can model the probability density of a molecular system \cite{kohler2020equivariant, wu2020stochastic, garcia2021n, DBLP:conf/iclr/XuLBPT21}, sample a lattice model \cite{li2018neural, albergo2019flow, nicoli2020asymptotically, nicoli2021estimation, boyda2021sampling}, or estimate free energies \cite{wirnsberger2020targeted, ding2021deepbar}. 
While a $\mathcal{C}^1$-diffeomorphic $T$ may be sufficient in a typical normalizing flow for %used to 
generating images or texts, %However, 
modeling physical systems requires more conditions such as a normalizing flow being a $\mathcal{C}^k$-diffeomorphism. For example, Boltzmann generator \cite{noe2019boltzmann,kohler2021smooth, liu2022path, ahmad2022free, jing2022torsional} %is an example 
requires the condition that $T$ is a $\mathcal{C}^2$-diffeomorphism. 

Boltzmann generators are generative models for sampling molecular structures whose existence probability distributions follow the Boltzmann distributions. Without loss of generality, %Ignoring the constants and dimension, 
the true target distribution can be expressed as $p^\star_{\mathbf{x}}(\mathbf{x}) \propto \exp{(-v(\mathbf{x}))}$ where $v$ is the potential energy of a molecular system and its %. Here, the 
force components are %defined as: % follows, 
\begin{equation}\label{eq1.2}
    \mathbf{f}(\mathbf{x}) =  - \partial_{\mathbf{x}} v(\mathbf{x}) = \partial_{\mathbf{x}} \log {p^\star_{\mathbf{x}}(\mathbf{x})} = \partial_{\mathbf{x}}p^\star_{\mathbf{x}}(\mathbf{x})/{p^\star_{\mathbf{x}}}(\mathbf{x}).
\end{equation}
Since valid molecular systems have well-defined and continuous force components, %. This fact and Equation \eqref{eq1.2} imply that 
$p^\star_{\mathbf{x}}$ must be at least once continuously differentiable (for continuous $\mathbf{f}$) and nonzero (for well-defined $\mathbf{f}$). 

When the Boltzmann generators are constructed using the normalizing flows, Equation \eqref{eq1.1} implies that $T^{-1}$ must be at least twice continuously differentiable (for continuously differentiable $p_{\mathbf{x}}$) and its derivative must have positive lower bound (for $p_{\mathbf{x}}$ to have positive lower-bound). In other words, $T$ should be at least $\mathcal{C}^2$-diffeomorphic. In this case, it is possible to train through force matching (FM) %is possible to 
by minimizing the mean squared error of the force components, which can improve the performance of Boltzmann generators. However, most existing normalizing flow models~\cite{DBLP:conf/iclr/DinhSB17, muller2019neural, durkan2019cubic, dolatabadi2020invertible, durkan2019neural} can not be used for this problem since they are only $\mathcal{C}^1$-diffeomorphic. They can not be trained using FM and can produce physically invalid samples. %. Not only can these models not be trained using FM, but they can also 
Smooth normalizing flows \cite{kohler2021smooth} addressed this problem by employing an ensemble of smooth ($\mathcal{C}^\infty$) bump functions \cite{Tu2008}. However, since those functions do not admit analytic inverses, either density estimation or sampling should be conducted in a time-consuming black box root finding method. To the best of the authors' knowledge, there is no known normalizing flow to meet the conditions of both being at least $\mathcal{C}^2$-diffeomorphic and admitting analytic inverses.
%, which is .

In this paper, we propose $\mathcal{C}^k$-diffeomorphic non-uniform B-spline flows where $k$ can be controlled so that they can be applicable to physics problems such as Boltzmann generators that require both at least $\mathcal{C}^2$-diffeomorphism and analytic inverses. %can be applied to  
We firstly investigate the sufficient condition for obtaining \emph{global invertibility}. To ensure that the probability density is continuously differentiable and has a positive lower bound, we state and prove the sufficient conditions for non-uniform B-spline transformations of any order to be bi-Lipschitz. Then, we propose a parameterization method that is applicable in periodic and aperiodic domains as a building block of our non-uniform B-spline flows to ensure %Moreover, to make  have
\emph{surjectivity} and \emph{sufficient expressive power}. %,  
Bi-Lipschitz continuity, surjectivity, and the nature of the non-uniform B-splines %, it is 
ensure that the proposed normalizing flow % transformation 
is a $\mathcal{C}^{k-2}$-diffeomorphism for the order $k$ non-uniform B-splines (using polynomials with degree $k-1$). %Through 
Our experiments %, we 
demonstrate that the proposed non-uniform B-spline flow is capable of solving % can also solve 
the FM problems in Boltzmann generators that can not be done with most existing normalizing flows and is able to solving the same problems quickly using analytic inverses for the low-order B-splines (\textit{i.e.}, polynomial root formula) that smooth normalizing flows can not admit. %. In addition, that only smooth normalizing flows can solve. Moreover, we also verify that low-order B-splines, which have an , can solve the same problem much faster.
Here is the summary on the contributions of this paper:
\begin{itemize}
    \item Investigating the sufficient conditions for the any order non-uniform B-spline transformation to be diffeomorphic on a compact domain.
    \item Proposing a parametrization method for the non-uniform B-spline transformation in normalizing flows to maintain expressive power on periodic / aperiodic domains.
    \item Showing that our non-uniform B-spline flows with FM yielded much better experimental results than RQ-spline flows and are %trained through force matching are 
    as good as the state-of-the-art smooth normalizing flows in density estimation and sampling.
    \item Demonstrating that sampling with non-uniform low-order B-spline flows is much faster than that of smooth flows due to the admissibility of analytic inverses.
\end{itemize}

\section{Related Works}

\paragraph{Coupling transformations.}
A coupling transform \cite{DBLP:journals/corr/DinhKB14, DBLP:conf/iclr/DinhSB17} $\phi : \Omega \rightarrow \Omega \subseteq \mathbb{R}^D$ is defined as
\begin{equation}\label{eq_flow}
    \phi(\mathbf{x})_i = \begin{cases}
    f_{\theta_i}(x_i), \theta_i = \mathrm{NN}(x_{1:d-1}) & \text{if } d \leq i \leq D, \\
    x_i & \text{if } 1 \leq i < d,
    \end{cases}
\end{equation}
where $\mathrm{NN}$ is an arbitrary neural network and $f_{\theta_i}:\Omega' \rightarrow \Omega' \subseteq \mathbb{R}$ is an invertible function parameterized by $\theta_i$.
The Jacobian determinant of this transform is easily obtained by the derivatives of $f$, expressed as $\mathrm{det} \left( {\partial \phi}/{\partial \mathbf{x}} \right) = \Pi_{i=d}^{D} {\partial f_{\theta_i}}/{\partial x_i}$.
The inverse of $\phi$ is obtained as
\begin{equation*}
    \phi^{-1}(\mathbf{x})_i = \begin{cases}
    f^{-1}_{\theta_i}(x_i), \theta_i = \mathrm{NN}(x_{1:d-1}) & \text{if } d \leq i \leq D, \\
    x_i & \text{if } 1 \leq i < d.
    \end{cases}
\end{equation*}
Coupling transformations have the computational advantages for %(i.e., 
Jacobian and inverse as well as %) yet 
have sufficient expressive power; hence many normalizing flows employ multiple coupling layers. However, various invertible $f$ functions have been proposed.
%different methods have designed various $f$. % can be used depending on the purpose.

\paragraph{Affine coupling flows.}
Many studies employ affine transformations as $f$, %al comfort, 
\textit{i.e.}, $f_{\theta_i}(x_i)=a_i x_i + b_i$ where $\theta_i = \{a_i, b_i\}$, for computational advantages. Since % Because 
affine transformations are easy to compute their Jacobian and inverse transformations~\cite{DBLP:journals/corr/KingmaW13, DBLP:conf/iclr/DinhSB17}, they are suitable for generating images \cite{kingma2018glow,ho2019flow++,lugmayr2020srflow,sukthanker2022generative} or speeches \cite{prenger2019waveglow,DBLP:conf/aaai/HeZ0LHY22} with %, which have 
relatively large dimensions $D$.
\begin{comment}
GLOW \cite{DBLP:conf/nips/KingmaD18},
SRflow \cite{DBLP:conf/eccv/LugmayrDGT20},
WaveGLOW \cite{DBLP:conf/icassp/PrengerVC19}
Flow++ \cite{pmlr-v97-ho19a}
ViTFlow \cite{DBLP:journals/corr/abs-2106-03959} (elementwise multiplication)
LipToSpeech \cite{DBLP:conf/aaai/HeZ0LHY22}
\end{comment}

\paragraph{Spline-based flows.}
Affine coupling flows have difficulty in learning discontinuous distributions even with small dimension $D$. % is small. 
To tackle this problem, some studies use more complex transformations such as splines,
%to deal with this problem. Splines are just 
which are piecewise polynomials or rational functions (parameterized with $\theta_i$), such as %Some examples are
linear and quadratic splines \cite{muller2019neural}, cubic splines \cite{durkan2019cubic}, linear-rational splines \cite{dolatabadi2020invertible}, and rational-quadratic (RQ) splines \cite{durkan2019neural}.
For the low-order splines with proper invertibility conditions,
%; therefore, they have the advantage that 
the inverse can be analytically obtained through the root formula, which is % (though 
slower than inverse affine transformations. %).

\paragraph{Smooth normalizing flows.}
Smooth normalizing flows \cite{kohler2021smooth} generate $\mathcal{C}^\infty$-diffeomorphism using smooth compact bump functions \cite{Tu2008} as follows: %, i.e., 
\begin{equation}
\begin{aligned}
    & f_{\theta_i}(x_i) = \sum_j \frac{\rho_{\theta_{ij}}(x_i)}{\rho_{\theta_{ij}}(x_i) +\rho_{\theta_{ij}}(1-x_i)},\\
    & \rho_{\theta_{ij}}(x)=\exp\left(-{1}/{(\alpha_{ij}x^{\beta_{ij}})}\right),     
\end{aligned}
\end{equation}
where $\theta_i=\bigcup_j \theta_{ij} = \bigcup_j \{\alpha_{ij},\beta_{ij}\}$.
Even though $\rho(x)$ is a low-order polynomial such as $x^3$ (which should have an analytic inverse), this function does not have %obviously has 
an analytic inverse %.  inverse of $f$ cannot be computed analytically 
because $f$ is an ensemble (\textit{i.e.}, linear combination) of rational functions.

\section{Diffeomorphic Non-uniform B-spline Flows}

Non-uniform B-splines \cite{curry1947spline, de1978practical} are highly attractive methods for compromising the trade-offs between spline spline-based flows and smooth normalizing flows. %' disadvantages.
Non-uniform B-splines have several nice properties: continuously differentiable with any degrees, %for any number of times, 
compact support, and locally analytic invertibility for low-orders. However, for %to actually build
constructing $\mathcal{C}^k$-diffeomorphic normalizing flows using non-uniform B-splines, the following conditions should be satisfied: global invertibility, surjectivity on various domains, and appropriate parameterization for sufficient expressive power.

\subsection{Definition for Flow Models}

We utilize the %introduce a 
definition of a non-uniform B-spline %given by Curry and Schoenberg 
in~\cite{curry1947spline} with some modifications on normalization suggested by %de Boor
\cite{de1978practical}. Let $t := \{t_j\}$ be an increasing sequence. The $j$-th non-uniform B-spline of order $k$ (using polynomials with degree $k-1$) for the knot sequence $t$ is denoted by $B_{j,k,t}$ and is %can be 
defined recursively as:
\begin{equation}\label{eq3.1.1}
    B_{j,1,t} = \mathbf{1}_{[t_j,t_{j+1})},
\end{equation}
\begin{equation}\label{eq3.1.2}
    B_{j,k,t} = \omega_{jk}B_{j,k-1,t} + (1 - \omega_{j+1,k}) B_{j+1,k-1,t},
\end{equation}
with
\begin{equation}\label{eq3.1.3}
    \omega_{jk}(x) := {(x-t_j)}/{(t_{j+k-1}-t_j)}.
\end{equation}
For %the sake of notational 
simplicity, we often drop $t$ so that $B_{j,k,t} = B_{jkt} = B_{jk}$. % or % so that $B_{jk} = B_{jkt}$.

Then,
%Referring to the flow model in the related work and the definition of non-uniform B-spline, 
for the flow model in Equation (\ref{eq_flow}),
we propose to construct 
the transformation $f:\mathbb{R} \rightarrow \mathbb{R}$ using the non-uniform B-splines of order $k$ (\textit{i.e.}, $B_{jkt}$) as follows:
\begin{equation}\label{e3.1.4}
    f(x;\mathbf{\alpha},t) = \sum_{j=r-k+1}^{s-1}{\alpha_j B_{jkt}(x)}, \forall x \in [t_r,t_s]
\end{equation}
where $r,s\in \mathbb{Z}$, $\alpha = \{\alpha_j\}_{j=r-k+1}^{s-1} \in \mathbb{R}^{s-r+k-1}$ and $t = \{t_j\}_{j=r-k+2}^{s+k-2} \in \mathbb{R}^{s-r+k-3}$ are designed to be the outputs of an arbitrary neural network (NN) with some constraints that %, which 
we further discuss later.
The function $f$ is a surjective mapping from $[t_r,t_s]$ to $[t_r,t_s]$ and this transformation is defined to be an identity mapping when the input is outside this range. %, thus we take the transformer as the same as  
Note that $f(\cdot;\mathbf{\alpha},t) \in \mathcal{C}^{k-2}$, but there is no guarantee that it is diffeomorphic.

A differentiable transformation $f$ is %called a
diffeomorphic if it is bijective and the inverse $ f^{-1}$ is differentiable. % as well. 
If these functions are $n$ times continuously differentiable, $f$ is called a $\mathcal{C}^{n}$-diffeomorphism. For a diffeomorphic $f$,
%To make $f$ a diffeomorphism, 
$(f^{-1})'$ should exist (\textit{i.e.}, bounded and nonzero) and thus, %. That is, 
there should be a positive lower bound for $f'(x;\mathbf{\alpha},t)$. Since $(f^{-1}(x))' = 1/f'(f^{-1}(x))$, %there should exist 
both the lower and upper bounds for $f'$ and $(f^{-1})'$ should exist, respectively.
Therefore, to enforce %Hence, to make 
$f$ to be diffeomorphic, NN should generate $\mathbf{\alpha}$ and $t$ so that $f'(\cdot;\mathbf{\alpha},t)$ is bounded on both sides. 
Let $\{t_j\}_{j=r-k+2}^{s+k-2}$ be an increasing sequence and $u,l \in \mathbb{R}, u>1>l>0$. Let $S(t,l,u) \subset \mathbb{R}^{s-r+k-1}$ be the set of $\mathbf{\alpha}$ that makes the derivative of $f(\cdot;\mathbf{\alpha},t)$ has an upper bound $u$ and a lower bound $l$, which can be written as
\begin{equation*}
    S(t,l,u)=\{\mathbf{\alpha}\in\mathbb{R}^{s-r+k-1} : l < f'(x;\mathbf{\alpha},t) < u ,\forall x \in \mathbb{R} \}.
\end{equation*}
Since $f'(x;\mathbf{\alpha},t)$ is linear in $\mathbf{\alpha}$, $S(t,l,u)$ is (open) convex. 
%Also, $S$ is bounded obviously. Moreover, $\mathbf{0} \in S(t,l,u)$ because $f(x;\mathbf{0},t) = x$ strictly increases.

%In this section, we try to find $S$ to make our B-spline transformations diffeomorphic.
\subsection{Sufficient Conditions for Diffeomorphism}
%In this subsection, 
We investigate the sufficient condition for non-uniform B-spline transformations of any order to be diffeomorphic.
There have been studies on sufficient conditions for diffeomorphic uniform B-spline transformations %to be 
\cite{chun2009simple,sdika2013sharp}. We leverage these studies to further investigate the sufficient condition for diffeomorphic \emph{non-uniform} B-spline transformations.
%follow the method of proof of those studies.
%See the supplementary material 
%All proofs of the theorems are provided in the supplementary material.

\begin{thm}[Sufficient condition for diffeomorphic transformations]\label{theorem1}
Let $k \in \mathbb{N} \setminus\{1,2\}$, $r,s\in \mathbb{Z}$, $r+k \leq s$. Let $\alpha = \{\alpha_j\}_{j=r-k+1}^{s-1} \in \mathbb{R}^{s-r+k-1}$ and $t = \{t_j\}_{j=r-k+1}^{s+k-1} \in \mathbb{R}^{s-r+2k-1}$ be (strictly) increasing sequences.
Let $f(x;\mathbf{\alpha},t) = \sum_{j=r-k+1}^{s-1}{\alpha_j B_{jkt}(x)}$, where $B_{jkt}$ is the $j$th non-uniform B-spline of order $k$ (from polynomials with degree $k-1$) for the knot sequence $t$. For $u>1$, $0<l<1$ and $j = r-k+2,\dots,s-1$,
\begin{equation} \label{eq_suff}
    \frac{l}{k-1} <\frac{\alpha_j - \alpha_{j-1}}{t_{j+k-1} - t_{j}} < \frac{u}{k-1}
\end{equation}
%for  then 
leads to $l < f'(\cdot;\mathbf{\alpha},t) < u$ on $[t_{r},t_{s}]$.
\end{thm}
\begin{proof}
By properties of the non-uniform B-spline \cite{de1978practical}, we have
\begin{equation}
\begin{aligned}
    & D\left(\sum_{j=r-k+1}^{s-1}{\alpha_j B_{jk}}\right) \\
    & \quad = \sum_{j=r-k+2}^{s-1}{(k-1)\frac{\alpha_j - \alpha_{j-1}}{t_{j+k-1} - t_{j}} B_{j,k-1}} \quad \text{on} \quad  [t_{r},t_{s}],
\end{aligned}
\end{equation}
\begin{equation}
    \sum_{j=r-k+2}^{s-1}{B_{j,k-1}} = 1 \quad \text{on} \quad  [t_{r+1},t_{s}],
\end{equation}
and
\begin{equation}
    \sum_{j=r-k+2}^{s-1}{B_{j,k-1}} < 1 \quad \text{on} \quad  [t_{r},t_{r+1}].
\end{equation}

Using above properties, we have
\begin{comment}
\begin{equation}
\begin{aligned}
    & \frac{l}{k-1} <\frac{\alpha_j - \alpha_{j-1}}{t_{j+k-1} - t_{j}} < \frac{u}{k-1} \forall j \in [r,s+1] \\
    \Rightarrow & \sum_{j=r}^{s+1}{(l)B_{j,k-1}} < D\left(\sum_{j=r}^{s}{\alpha_j B_{jk}}\right) < \sum_{j=r}^{s+1}{(u)B_{j,k-1}}\\
    \Rightarrow & \sum_{j=r}^{s+1}{(l)B_{j,k-1}} < D\left(\sum_{j=r}^{s}{\alpha_j B_{jk}}\right) < \sum_{j=r}^{s+1}{(u)B_{j,k-1}},
\end{aligned}
\end{equation}
\end{comment}
\begin{enumerate}
\item[] $\frac{l}{k-1} <\frac{\alpha_j - \alpha_{j-1}}{t_{j+k-1} - t_{j}} < \frac{u}{k-1}$ for $j = r-k+2,\dots,s-1$
\item[$\Rightarrow$] $ \sum_{j=r-k+2}^{s-1}{lB_{j,k-1}} < D\left(\sum_{j=r-k+1}^{s-1}{\alpha_j B_{jk}}\right) < \sum_{j=r-k+2}^{s-1}{uB_{j,k-1}}$
\item[$\Rightarrow$] $l\sum_{j=r-k+2}^{s-1}{B_{j,k-1}} < D\left(\sum_{j=r-k+1}^{s-1}{\alpha_j B_{jk}}\right) < u\sum_{j=r-k+2}^{s-1}{B_{j,k-1}}$
\item[$\Rightarrow$] $l < D\left(\sum_{j=r-k+1}^{s-1}{\alpha_j B_{jk}}\right) < u$ \quad on \quad $[t_{r},t_{s}]$
\item[$\Rightarrow$] $l < f'(\cdot;\mathbf{\alpha},t) < u$ \quad on \quad $[t_{r},t_{s}]$.
\end{enumerate}
\end{proof}

Theorem \ref{theorem1} suggests that we can generate bi-Lipschitz non-uniform B-spline transformations by constraining %using 
the parameters $\alpha = \{\alpha_j\}_{j=r-k+1}^{s-1}$and $t = \{t_j\}_{j=r-k+1}^{s+k-1}$ to satisfy (\ref{eq_suff}). Then, by the nature of non-uniform B-splines, we can guarantee that these bi-Lipschitz $k$th-order non-uniform B-spline transformations are in fact $\mathcal{C}^{k-2}$-diffeomorphic.

\subsection{Existence of an Analytic Inverse}

Smooth normalizing flows exhibit good expressive power but with the price of slow non-analytic inverse transformations. In contrast, the non-uniform B-splines of order $k$ have analytic inverses % transformations 
if $k < 5$ %is less than 5 
since they are piecewise $(k-1)$th-order polynomials. %, non-uniform B-splines 
Inverse non-uniform B-spline transformation has been partially studied in \cite{tristan2007fast}, which %but it 
ensured its exactness only on the knots $t$, not on the entire continuous domain. Here, we propose an analytic inverse non-uniform B-spline transformation for Equation \eqref{e3.1.4}
%\eqref{eq3.1.1} 
on a compact domain %so that it can be used 
in our normalizing flows with $k < 5$.

The map $f:[t_r,t_s]\rightarrow[t_r,t_s]$ in Equation \eqref{e3.1.4} with $\alpha \in \mathbb{R}^{s-r+k-1}$ satisfying the sufficient condition \eqref{eq_suff} in Theorem \ref{theorem1} is %we can generate a 
$\mathcal{C}^{k-2}$-diffeomorphic. %  as in .
%$\mathcal{C}^{k-1}$-diffeomorphic. %  as in .
Like prior works \cite{durkan2019neural,durkan2019cubic}, computing the inverse of a non-uniform B-spline at any location $y$ requires finding the bin index $j \in [r,s]$ where $x$ lies with %, \textit{i.e.}, %,  one needs to find the  such that 
$ f(t_j;\alpha, t) \leq y < f(t_{j+1};\alpha, t)$. %By using t
The following equation can easily identify %, we can find 
such bin $j$: % easily.
\begin{equation*}%\label{eq3.3.1}
    f(t_j;\mathbf{\alpha},t) = \sum_{i=r-k+1}^{s-1}{\alpha_i B_{ik} (t_j)}
    = \sum_{i=j-k+1}^{j-1}{\alpha_i B_{ik} (t_j)}.
\end{equation*}
This bin search %Note that this 
does not increase computational burden due to %since 
strictly increasing (\textit{i.e.}, sorted) $\{f(t_j;\mathbf{\alpha},t)\}_j$ %.  is a  
sequence.

After finding the bin $j$, %we know that 
$\{i | \text{supp}(B_{ik}) \cap (t_j,t_{j+1}) \neq \emptyset \} = \{j-k+1, j-k+2, \dots, j\}$. Thus, %we have
for the given $y$, the following equation holds for $x \in [t_j,t_{j+1})$,
\begin{equation}\label{eq3.3.2}
    y = \sum_{i=j-k+1}^{j}{\alpha_i B_{ik} (x)}.
\end{equation}
Equation \eqref{eq3.3.2} is a $(k-1)$th-order polynomial, so % equation. Hence we can use 
the root formulas of the polynomial equations can be used if $k \leq 5$. In case of $k=4$ (\textit{i.e.}, cubic B-spline), the %we use a 
root finding algorithm %provided 
by \cite{Peters2016How} can be used, which is a modification of the algorithm of \cite{blinn2007solve}. 
We used the code of %borrow some of the official source code for 
cubic-root computation %calculation created 
by \cite{durkan2019cubic}.

\subsection{Definitions on Various Domains}
For the applications to physics problems, non-uniform B-spline transformation must be well-defined on domains such as %in both domains, which are 
closed interval and circle (\textit{i.e.}, periodic interval). Without loss of generality, %If we can build
we construct transformations on $\mathbb{I}$ (unit interval) and $S^1$ (unit circle), which can be extended to
%, we can also build transformations on 
arbitrary closed intervals and circles through affine transforms.
The following subsections discuss how to construct
%This subsection discusses how to build 
well-defined diffeomorphic non-uniform B-spline transformations on $\mathbb{I}$ and $S^1$. 

\paragraph{On $\mathbb{I}$.}
Two constraints $f(0) = 0$ and $f(1) = 1$ must hold to make $f$ be surjective from $\mathbb{I}$ to $\mathbb{I}$.
One trivial solution for a surjective $f$ (\textit{i.e.}, $f(0) = 0$ and $f(1) = 1$) is to %If we na\"ively 
set $\alpha_j = 0$ for all $j$ such that $t_j<0$ and $\alpha_j = 1$ for all $j$ such that $t_{j+k}>1$. %, then the surjectivity  holds.
However, this na\"ive solution severely decreases the expressive power of the non-uniform B-spline transformation near both endpoints (\textit{i.e.}, 0 and 1). This is because the $k$th-order non-uniform B-spline transformations are $\mathcal{C}^{k-2}$, which results in $f^{(m)}(0)=0$ and $f^{(m)}(1)=0$ for $m = 1,\dots,k-2$ where %. Note that 
$f^{(m)}(x)$ denotes the $m$th derivative of $f(x)$. 
Therefore, we propose Algorithm \ref{alg:algorithm} to generate parameters (\textit{i.e.}, $t$ and $\alpha$) %as in  
to maintain the expressive power of the non-uniform B-spline transformation at both endpoints.
This proposed algorithm consists of three main steps as follows:
\begin{itemize}
\item{Line 1-7} : Generating an increasing sequence $\{t_j\}_{j=r-k+2}^{s+k-2}$ such that $t_r=0$, $t_s=1$ and $\Delta t_j := t_{j+1} - t_j$ having a positive infimum for $j=r-k+2, \dots, s+k-3$. Such a sequence can be obtained by applying the softmax (line 1), cumulative summation (line 4-7), and the affine transform (line 2-3) to the output of an arbitrary neural network. Note that $\epsilon_t$ is set to a small positive constant (\textit{e.g.}, $10^{-6}$) to ensure $\Delta t_j$ has a positive infimum.
\item{Line 8-12} : Similar to the first step, generating another increasing sequence $\{\alpha_j\}_{j=r-k+2}^{s-2}$ such that %, subject to 
$\alpha_{r-k+2}>0$, $\alpha_{s-2}<1$ and $\Delta \alpha_j := \alpha_{j+1} - \alpha_j$ having a positive infimum for $j=r-k+2, \dots, s-3$. $\epsilon_\alpha$ is set to a small positive constant (\textit{e.g.}, $10^{-6}$) to ensure $\Delta \alpha_j$ has a positive infimum.
\item{Line 13-17} : Computing $\alpha_{r-k+1}$ and $\alpha_{s-1}$ such that $f(0)=0$ and $f(1)=1$, respectively. This step ensures surjectivity.
\end{itemize}

\begin{algorithm}[tb]
\caption{Non-uniform B-spline parameter generation}
\label{alg:algorithm}
\textbf{Input}: $\Delta \Tilde{t} = \left\{\Delta \Tilde{t}_j\right\}_{j=r-k+2}^{s+k-3}, \Delta \Tilde{\alpha} = \left\{\Delta \Tilde{\alpha}_j\right\}_{j=r-k+1}^{s-2}$\\
\textbf{Parameter}: $\epsilon_t, \epsilon_\alpha$\\
\textbf{Output}: $t = \left\{t_j\right\}_{j=r-k+2}^{s+k-2}, \alpha = \left\{\alpha_j\right\}_{j=r-k+1}^{s-1}$\\
\begin{algorithmic}[1] %[1] enables line number
\STATE $\Delta t \leftarrow \mathrm{softmax}(\Delta \Tilde{t})$
\STATE $\Delta t \leftarrow \epsilon_t + (1 - (s-r+2k-4)\epsilon_t)\Delta t$
\STATE $\Delta t \leftarrow \Delta t / \sum_{j=r}^{s-1}(\Delta t)_j$
\STATE $t_{r-k+2} = - \sum_{j=r-k+2}^{r-1} \Delta t_j$
\FOR{$i = r-k+3,\dots,s+k-2$}
\STATE $t_i=t_{r-k+2}+\sum_{j=r-k+2}^{i-1} \Delta t_j $
\ENDFOR
\STATE $\Delta \alpha \leftarrow \mathrm{softmax}(\Delta \Tilde{\alpha})$
\STATE $\Delta \alpha \leftarrow \epsilon_\alpha + (1 - (s-r+k-2)\epsilon_\alpha)\Delta \alpha$
\FOR{$i = r-k+2,\dots,s-2$}
\STATE $\Tilde{\alpha}_i=\sum_{j=r-k+1}^{i-1} \Delta \alpha_j $
\ENDFOR
\STATE $\Tilde{\alpha}_{r-k+1}=0$
\STATE $\Tilde{\alpha}_{s-1}=1$
\STATE $\Tilde{f_r}=\sum_{j=r-k+1}^{r-1} \Tilde{\alpha}_j B_{jkt}(t_r)$
\STATE $\Tilde{f_s}=\sum_{j=s-k+1}^{s-1} \Tilde{\alpha}_j B_{jkt}(t_s)$
\STATE $\alpha = (\Tilde{\alpha} -\Tilde{f_r}) / (\Tilde{f_s} - \Tilde{f_r})$
\STATE \textbf{return} $t,\alpha$
\end{algorithmic}
\end{algorithm}

\begin{comment}
Theorem \ref{theorem2} suggests that if we generate $\nabla \alpha = \{\nabla \alpha_j \}_{j=r-k+2}^{s-1}$ subject to Equation \eqref{e40} and Theorem \ref{theorem1}, $f$ becomes diffeomorphism from $[t_r,t_s]$ to $[t_r,t_s]$, whose derivative is bigger than $l$ and smaller than $u$. 
- need to connect this with (1), there is no $\Omega \subset \mathbb{R}$ defined in (1).

Any $\mathcal{C}^{n}$ diffeomorphism on an arbitrary interval $\Omega \subset \mathbb{R}$ can be implemented by scaling and shifting the $\mathcal{C}^{n}$ diffeomorphism on $\mathbb{I}$. Similarly, any $\mathcal{C}^{n}$ periodic diffeomorphism on an arbitrary closed interval $\Omega \subset \mathbb{R}$ can be implemented by re-scaling the $\mathcal{C}^{n}$ diffeomorphism on $S^1$. Therefore, this subsection demonstrates how to achieve diffeomorphic non-uniform B-spline transformations on the unit circle $S^1$ and the unit interval $\mathbb{I}$.
\end{comment}

\paragraph{On $S^1$.}
Smooth normalizing flows implemented periodic transformations with non-zero derivatives in interval boundaries. Since the vanishing gradient at the endpoint makes the universal approximation of arbitrary periodic transformations impossible, it is important to implement periodic transformations. Similar to RQ circular spline flows \cite{durkan2019neural}, we propose to control the parameters to match all $1$st to $(k-2)$th-order derivatives on both interval boundaries using the following theorem.

\begin{thm}\label{theorem2}
Let $k \in \mathbb{N}\setminus\{1,2\}$, $r,s\in \mathbb{Z}$, $r+k \leq s$. Let $\alpha = \{\alpha_j\}_{j=r-k+1}^{s-1} \in \mathbb{R}^{s-r+k-1}$ and $t = \{t_j\}_{j=r-k+1}^{s+k-1} \in \mathbb{R}^{s-r+2k-1}$ be (strictly) increasing sequences.
Let $f(x;\mathbf{\alpha},t) = \sum_{j=r-k+1}^{s-1}{\alpha_j B_{jkt}(x)}$, where $B_{jkt}$ is the $j$th non-uniform B-spline of order $k$ for the knot sequence $t$.
If $\alpha_{s-i} = 1 + \alpha_{r-i}$ for $i=1,2,\dots,k-1$ and $t_{s+j} = 1 + t_{r+j}$ for $j=-k+2,-k+3,\dots,k-3,k-2$, then $f^{(m)}(t_r) = f^{(m)}(t_s)$ for $m=1,\dots,k-2$.
\end{thm}
\begin{proof}
Non-uniform B-splines have the following property. \cite{de1978practical}
\begin{equation}
    D^m\left(\sum_{j}\alpha_j B_{jk} \right) = \sum_{j}\alpha_j^{(m+1)} B_{j,k-m}
\end{equation}
with
\begin{equation}
    \alpha_j^{(m+1)} = \begin{cases}
    \alpha_j, & \text{if } m=0, \\
    \frac{\alpha_j^{(m)}-\alpha_{j-1}^{(m)}}{(t_{j+k-m}-t_j)/(k-m)} & \text{if } m > 0,
    \end{cases}    
\end{equation}
where $D$ is the differential operator.
Using this property, we prove Theorem 2 by mathematical induction.

If $m=1$,
\begin{equation}
\begin{aligned}
    \alpha_{s-i}^{(2)} & = \frac{\alpha^{(1)}_{s-i}-\alpha^{(1)}_{s-i-1}}{(t_{s-i+k-1}-t_{s-i})/(k-1)}\\
    & = \frac{\alpha_{s-i}-\alpha_{s-i-1}}{(t_{s-i+k-1}-t_{s-i})/(k-1)}\\
    & = \frac{\alpha_{r-i}-\alpha_{r-i-1}}{(t_{r-i+k-1}-t_{r-i})/(k-1)} = \alpha_{r-i}^{(2)}
\end{aligned}    
\end{equation}
for $i=1,\dots,k-2$. And
\begin{equation}
\begin{aligned}
     D\left(\sum_{j}\alpha_j B_{jk} \right)(t_s) & = \sum_{j=s-k+1}^{s-1}\alpha_j^{(2)} B_{j,k-1}(t_s)\\
     & = \sum_{j=s-k+2}^{s-1}\alpha_j^{(2)} B_{j,k-1}(t_s)
\end{aligned}
\end{equation}
because $B_{s-k+1,k-1}(t_s) = 0$.
In the same way, we also have 
\begin{equation}
     D\left(\sum_{j}\alpha_j B_{jk} \right)(t_r) = \sum_{j=r-k+2}^{r-1}\alpha_j^{(2)} B_{j,k-1}(t_r).    
\end{equation}
By the assumption and Lemma 1, 
\begin{equation}
    \sum_{j=s-k+2}^{s-1}\alpha_j^{(2)} B_{j,k-1}(t_s) = \sum_{j=r-k+2}^{r-1}\alpha_j^{(2)} B_{j,k-1}(t_r).
\end{equation}
Therefore,
\begin{equation}
    D\left(\sum_{j}\alpha_j B_{jk} \right)(t_s) = D\left(\sum_{j}\alpha_j B_{jk} \right)(t_r) .
\end{equation}

Assume $m \leq k-3$, $\alpha_{s-i}^{(m+1)} = \alpha_{r-i}^{(m+1)}$ for $i=1,\dots,k-(m+1)$, and 
\begin{equation}
    D^m\left(\sum_{j}\alpha_j B_{jk} \right)(t_s) = D^m\left(\sum_{j}\alpha_j B_{jk} \right)(t_r). 
\end{equation}
Then
\begin{equation}
\begin{aligned}
    \alpha_{s-i}^{(m+2)} & = \frac{\alpha^{(m+1)}_{s-i}-\alpha^{(m+1)}_{s-i-1}}{(t_{s-i+k-(m+1)}-t_{s-i})/(k-(m+1))}\\
    & =\frac{\alpha^{(m+1)}_{r-i}-\alpha^{(m+1)}_{r-i-1}}{(t_{r-i+k-(m+1)}-t_{r-i})/(k-(m+1))}\\
    & = \alpha_{r-i}^{(m+2)}
\end{aligned}    
\end{equation}
for $i=1,\dots,k-(m+2)$.
Then
\begin{equation}
\begin{aligned}
     & D^{m+1}\left(\sum_{j}\alpha_j B_{jk} \right)(t_s) \\
     & \quad = \sum_{j=s-k+1}^{s-1}\alpha_j^{(m+2)} B_{j,k-(m+1)}(t_s)\\
     & \quad = \sum_{j=s-k+1+(m+1)}^{s-1}\alpha_j^{(m+2)} B_{j,k-(m+1)}(t_s),
\end{aligned}
\end{equation}
Because $B_{j,k-(m+1)}(t_s) = 0 $ for $j < s-k+1+(m+1)$.
In the same way, we have
\begin{equation}
\begin{aligned}
     & D^{m+1}\left(\sum_{j}\alpha_j B_{jk} \right)(t_r) \\
     & \quad = \sum_{j=r-k+1+(m+1)}^{r-1}\alpha_j^{(m+2)} B_{j,k-(m+1)}(t_r).
\end{aligned}
\end{equation}
By the assumption and Lemma 1, we get
\begin{equation}
    D^{m+1}\left(\sum_{j}\alpha_j B_{jk} \right)(t_s) = D^{m+1}\left(\sum_{j}\alpha_j B_{jk} \right)(t_r). 
\end{equation}
Therefore, $f^{(m)}(t_r) = f^{(m)}(t_s)$ for $m=1,\dots,k-2$.
\end{proof}

Theorem \ref{theorem2} suggests that by %if we 
taking additional conditions on some of the parameters, % to be the same, we can build
$\mathcal{C}^{k-2}$-diffeomorphic non-uniform B-spline transformations can be constructed on the unit circle.
Therefore, when we generate diffeomorphic non-uniform B-spline transformations on $S^1$, we use the Algorithm \ref{alg:algorithm}, but %as in $\mathbb{I}$ 
with $\alpha_{s-i} = 1 + \alpha_{r-i}$ for $i=1,2,\dots,k-1$ and $t_{s+j} = 1 + t_{r+j}$ for $j=-k+2,-k+3,\dots,k-3,k-2$.
%Therefore, we use the same algorithm when we generate diffeomorphic non-uniform B-spline transformations on $S^1$ as we do on $\mathbb{I}$.
These additional conditions can be implemented by setting 
$\Delta \Tilde{\alpha}_{s-i} = 1 + \Delta \Tilde{\alpha}_{r-i}$ for $i=2,\dots,k-1$ and $\Delta \Tilde{t}_{s+j} = 1 + \Delta \Tilde{t}_{r+j}$ for $j=-k+2,-k+3,\dots,k-3$ in the %To implement this condition on 
Algorithm \ref{alg:algorithm}. %, we just set 

\section{Experiments}

\subsection{Illustrative Toy Example}
\begin{figure}[tb]%
\centering
\begin{tabular}{ccc}
\begin{tabular}[c]{@{}c@{}}(a) Ground\\ Truth\end{tabular}                                                   & \multicolumn{2}{c}{\adjustbox{valign=m}{{\includegraphics[width=0.6\columnwidth ]{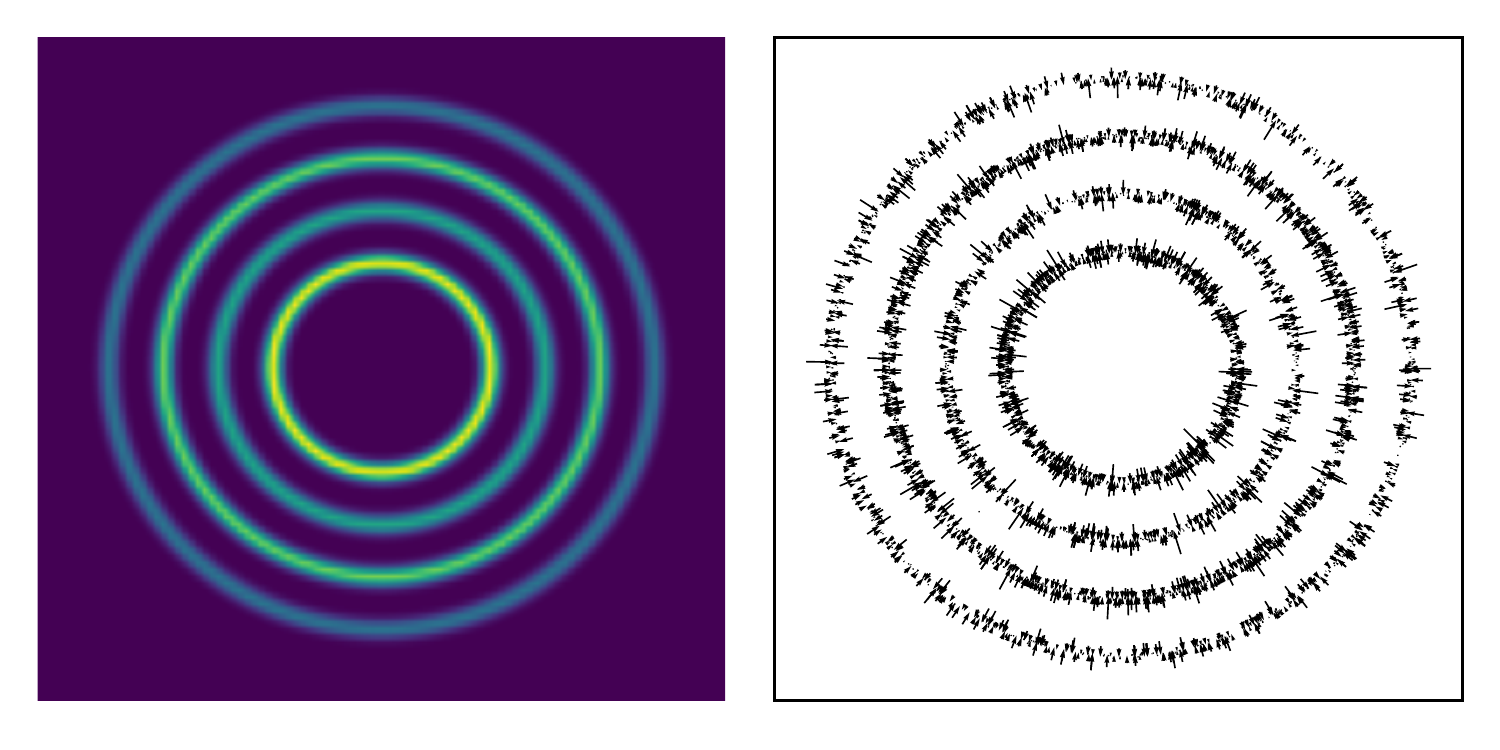} }}} \\
\begin{tabular}[c]{@{}c@{}}(b) RQ-\\spline\end{tabular}                                                      & \multicolumn{2}{c}{\adjustbox{valign=m}{{\includegraphics[width=0.6\columnwidth ]{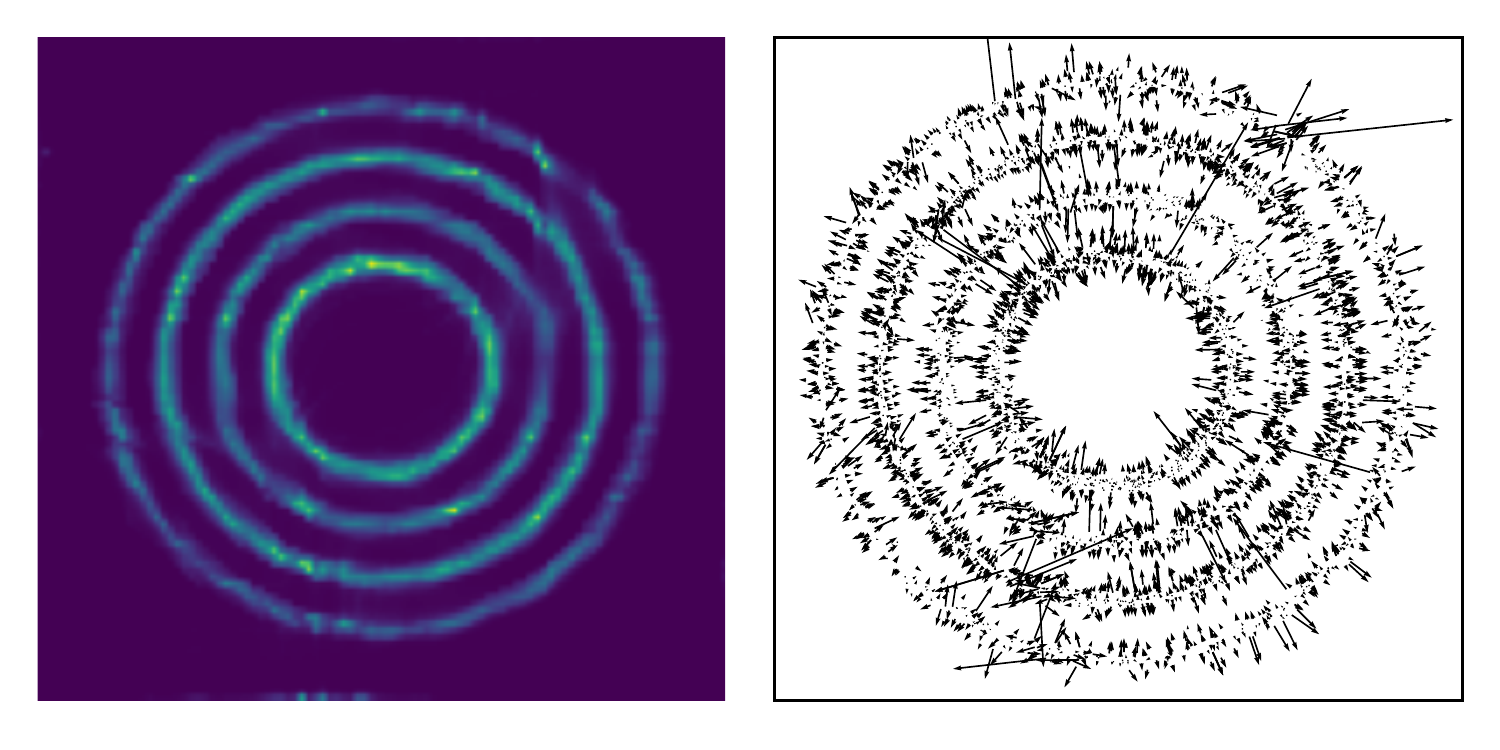} }}} \\
(c) Smooth                                                         & \multicolumn{2}{c}{\adjustbox{valign=m}{{\includegraphics[width=0.6\columnwidth ]{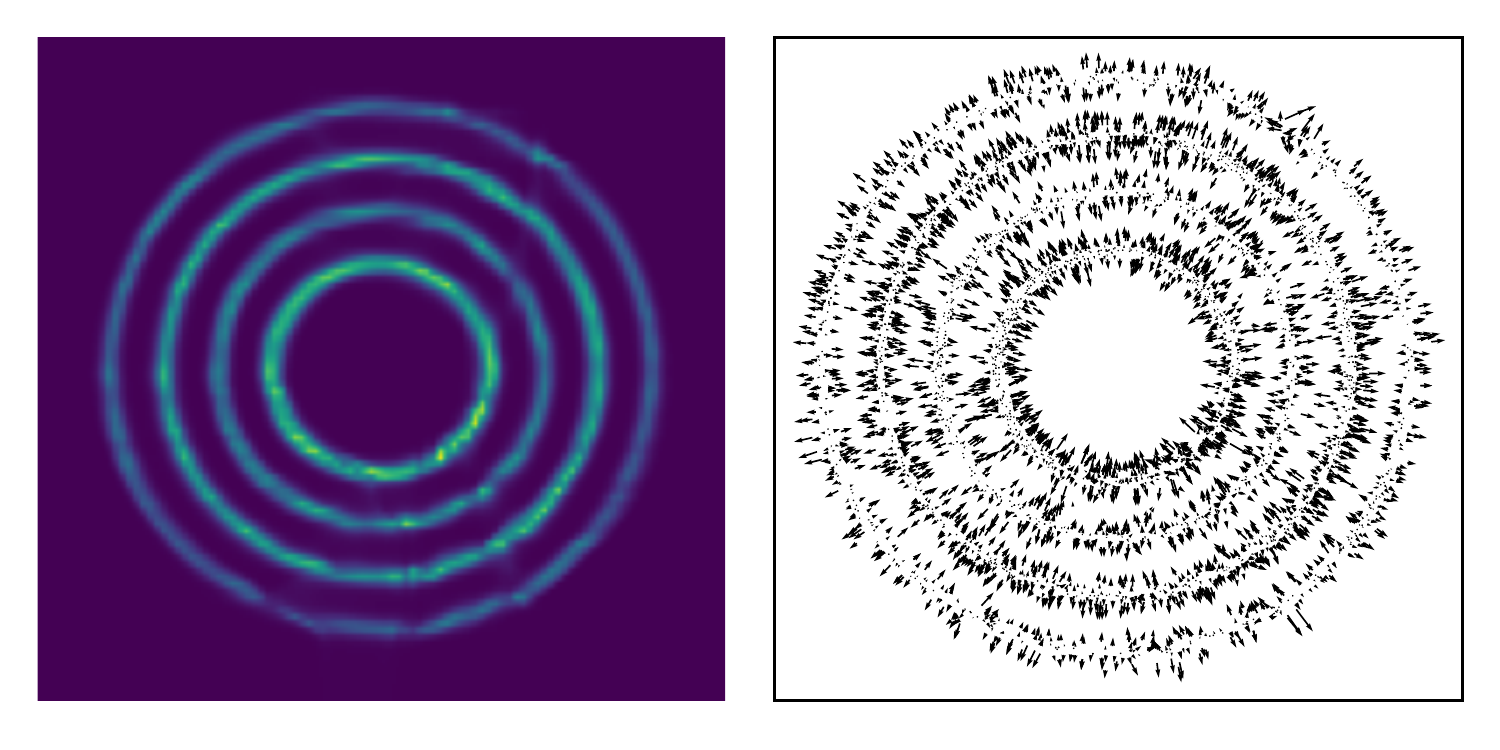} }}} \\
\begin{tabular}[c]{@{}c@{}}(d) Non-\\uniform\\ B-spline \\(ours)\end{tabular} & \multicolumn{2}{c}{\adjustbox{valign=m}{{\includegraphics[width=0.6\columnwidth ]{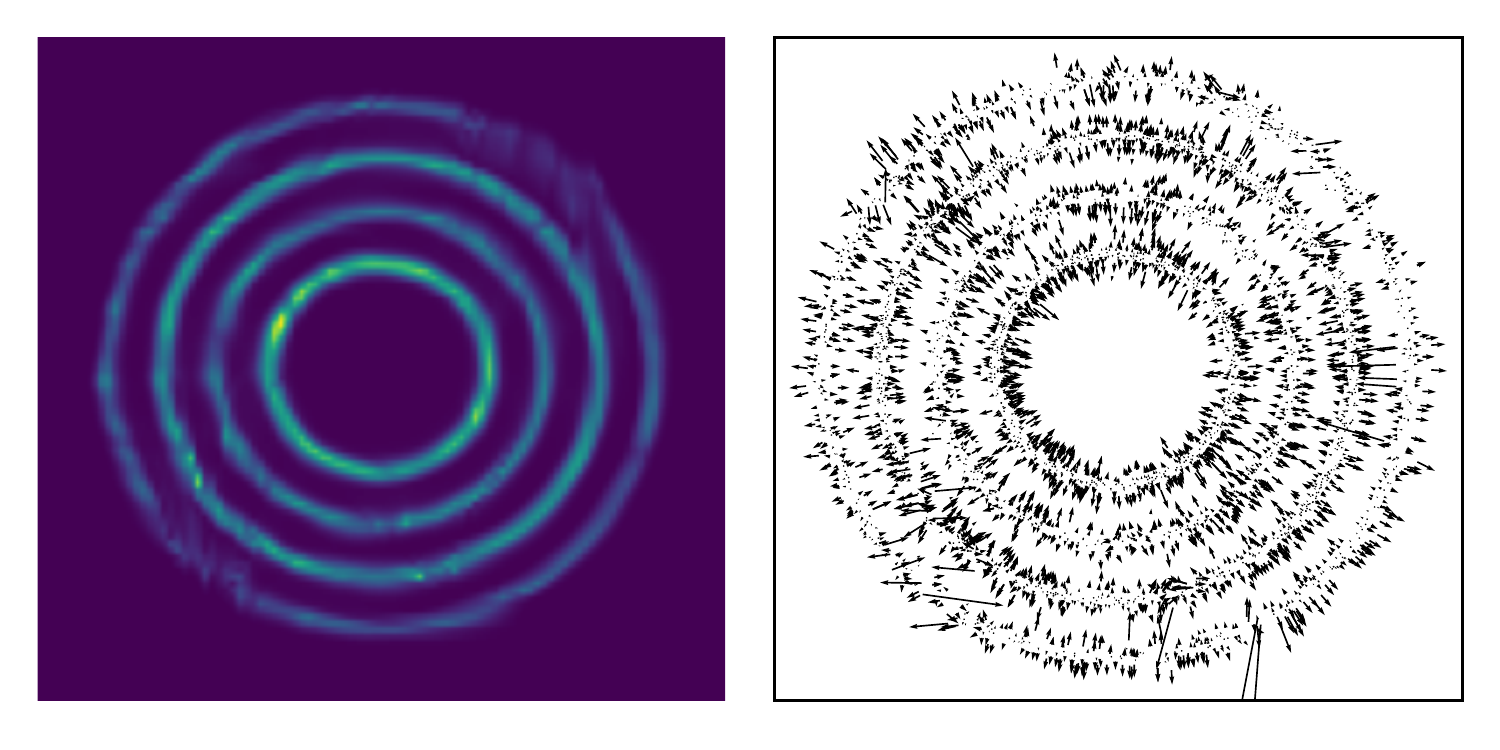} }}}\\
\end{tabular}
\caption{Probability density (left) and its corresponding force field (right) of (a) ground truth, and their approximations %approximated 
by (b) RQ-spline flows, (c) Smooth flows, and (d) Non-uniform (cubic) B-spline flows (ours). }%
\label{fig:toy}% 
\end{figure}

A simple toy example in this section shows that the proposed non-uniform B-spline flows generate feasible continuous forces. All flow models used NLL as a loss function. We provide full experimental details in the supplementary material. Figure \ref{fig:toy} (a) shows energy and force by two-dimensional ground-truth distribution. In Figure \ref{fig:toy} (b), RQ-spline flow shows an unstable force field with many singularities (\textit{i.e.}, unrealistic) because it is only a $\mathcal{C}^{1}$-diffeomorphism. On the other hand, since smooth normalizing flow is a $\mathcal{C}^{\infty}$-diffeomorphism, the force field is well-defined (\textit{i.e.}, not diverging) and continuous in Figure \ref{fig:toy} (c). In Figure \ref{fig:toy} (d), non-uniform cubic B-spline flow, which is a $\mathcal{C}^{2}$-diffeomorphism, shows that the force field is well-defined and continuous, which is comparable to %just like 
the smooth normalizing flow.

% do not use trim, clip
\begin{figure}[tb]% 
\subfloat[Aperiodic ($\mathbb{I}$)]{{\includegraphics[width=0.48\columnwidth ]{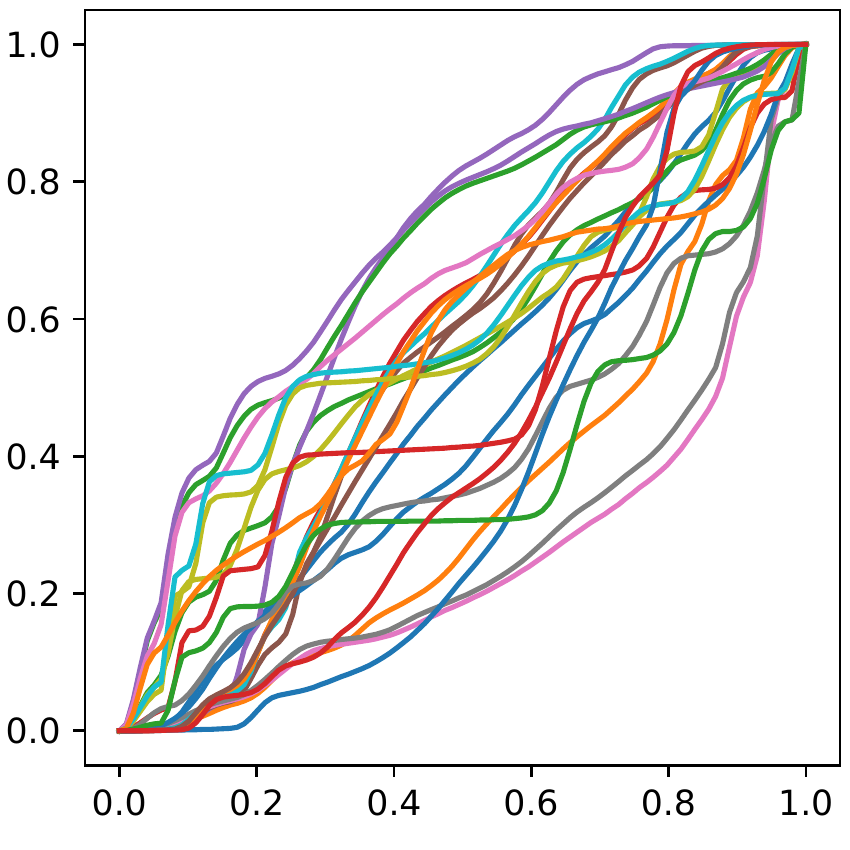} }}
\subfloat[Periodic ($\mathcal{S}^1$)]{{\includegraphics[width=0.48\columnwidth ]{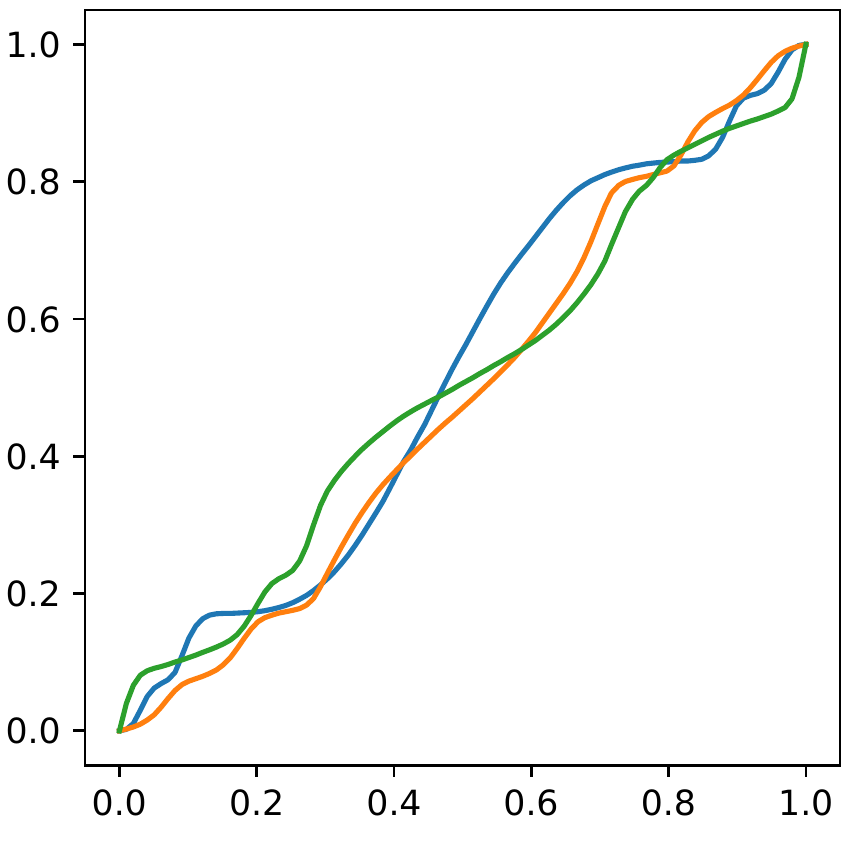} }}
\caption{Accumulation of non-uniform B-spline transformations on (a) $\mathbb{I}$ and (b) $\mathcal{S}^1$ generated by the flow model used in the illustrative toy example experiment ((a) from Figure~\ref{fig:toy}(d) and (b) from the periodic case %Figure \ref{fig:toy_periodic} 
in supplemental, respectively). The displayed transformations were randomly selected. %sampled during the experiment. 
Note that %we plotted much fewer
learned transformations in (b) have %to check out the
continuous derivatives %are continuous 
at both ends, demonstrating % (as in 
Theorem 2. }%
\label{fig:toy_accumulation}% 
\end{figure}

We also depict some non-uniform B-spline transformations generated by our models %used 
in this toy example experiment in Figure \ref{fig:toy_accumulation}. 
Figure \ref{fig:toy_accumulation} (a) shows the transformation in the aperiodic domain ($\mathbb{I}$), and Figure \ref{fig:toy_accumulation} (b) illustrates the transformation in the periodic domain ($\mathcal{S}^1$). Note that Figure \ref{fig:toy_accumulation} (a) is extracted from the flow that models aperiodic probability density as in Figure \ref{fig:toy}, while Figure \ref{fig:toy_accumulation} (b) is obtained from the flow that models periodic probability density as in the periodic toy example %Figure \ref{fig:toy_periodic} 
in supplemental. We can observe that our diffeomorphic non-uniform B-spline transformation is surjective, differentiable, and diversely expressive. %while has diverse expressivity. 
In particular, in Figure \ref{fig:toy_accumulation} (b), it can be observed that it also has the same slope at both endpoints as Theorem \ref{theorem2} guarantees while having diverse diffeomorphic transformations.

\subsection{Boltzmann Generator Training by Force Matching}
This section demonstrates that the proposed non-uniform B-spline flows can be applied to physical system modeling such asa Boltzmann generator, trained with FM, just like smooth normalizing flows, but unlike other prior normalizing flows such as RQ spline flows.
%, as well as smooth normalizing flows, can be applied to physical system modeling. We train a Boltzmann generator through FM. 
The output of the Boltzmann generator is a 60-dimensional vector representing the structure (including bond length and angle) of a molecular system, alanine dipeptide. Alanine dipeptide is a typical example of molecular system structure sampling problems. Because alanine dipeptide has a highly nonlinear potential energy surface and singular points, common $\mathcal{C}^1$-diffeomorphic normalizing flows may have difficulty in learning its distribution model.

We trained three normalizing flow models: RQ-spline flow, smooth normalizing flow, and our non-uniform B-spline flow. Two loss functions were used: % was twofold:
NLL loss $\mathcal{L}_{\text{NLL}}$ or NLL + FM loss $(1-\lambda_{\text{FM}})\mathcal{L}_{\text{NLL}} + \lambda_{\text{FM}} \mathcal{L}_{\text{FM}}$ where
\begin{equation}
\begin{aligned}
    \mathcal{L}_{\text{FM}} & = \frac{1}{N} \sum_{n=1}^{N} \left\lVert \mathbf{f}(\mathbf{x}^{(n)}) -  \partial_{\mathbf{x}} \log {p_{\mathbf{x}}(\mathbf{x}^{(n)})} \right\rVert_2^2 \\
    & \approx \mathbb{E}_{\mathbf{x} \sim p_{\mathbf{x}}^\star}\left[\left\lVert \mathbf{f}(\mathbf{x}) -  \partial_{\mathbf{x}} \log {p_{\mathbf{x}}(\mathbf{x})} \right\rVert_2^2 \right],     
\end{aligned}
\end{equation}
and we set $\lambda_{\text{FM}}=0.001$. In density estimation, each transformation is performed in a forward direction, and in sampling, it is performed in a reverse direction (\textit{i.e.}, root-finding). However, NLL + FM loss was not able to be used %available 
for RQ-spline flows since it diverged. % the loss did not even converge. 
So we trained Boltzmann generators in a total of five scenarios. Each experimental scenario was repeated ten times. Experimental details are described in the supplementary material.

\begin{table}[tb]
\small
\centering
\begin{tabular}{|l|c|c|c|}
\hline
Method                        & NLL                                                  & \begin{tabular}[c]{@{}c@{}}FME\\ $\times 10^4$\end{tabular}                                                    & KLD                                                  \\ \hline
RQ-spline                     & \begin{tabular}[c]{@{}c@{}}-210.28\\ ($\pm$ 0.05)\end{tabular} & \begin{tabular}[c]{@{}c@{}}79.95\\ ($\pm$ 92.80)\end{tabular} & \begin{tabular}[c]{@{}c@{}}385.22\\ ($\pm$ 8.05)\end{tabular} \\ \hline
Smooth                        & \begin{tabular}[c]{@{}c@{}}-210.87\\ ($\pm$ 0.05)\end{tabular} & \begin{tabular}[c]{@{}c@{}}1.34\\ ($\pm$ 0.07)\end{tabular} & \begin{tabular}[c]{@{}c@{}}192.57\\ ($\pm$ 0.22)\end{tabular} \\ \hline
Smooth+FM                   & \begin{tabular}[c]{@{}c@{}}-210.25\\ ($\pm$ 0.09)\end{tabular} & \begin{tabular}[c]{@{}c@{}}0.909\\ ($\pm$ 0.002)\end{tabular} & \begin{tabular}[c]{@{}c@{}}196.85\\ ($\pm$ 1.02)\end{tabular} \\ \hline
\begin{tabular}[l]{@{}l@{}} Non-uniform \\B-spline (ours)\end{tabular}      & \begin{tabular}[c]{@{}c@{}}-211.03\\ ($\pm$ 0.03)\end{tabular} & \begin{tabular}[c]{@{}c@{}}4.59\\ ($\pm$ 5.11)\end{tabular} & \begin{tabular}[c]{@{}c@{}}192.62\\ ($\pm$ 0.31)\end{tabular} \\ \hline
\begin{tabular}[l]{@{}l@{}}Non-uniform B-\\spline+FM (ours)\end{tabular} & \begin{tabular}[c]{@{}c@{}}-209.45\\ ($\pm$ 1.78)\end{tabular} & \begin{tabular}[c]{@{}c@{}}0.812\\ ($\pm$ 0.345)\end{tabular} & \begin{tabular}[c]{@{}c@{}}228.09\\ ($\pm$ 54.6)\end{tabular} \\ \hline
\end{tabular}
\caption{Negative log-likelihoods (NLLs), force matching errors (FMEs), and reverse Kullback-Leibler divergences (KLDs) of alanine dipeptide training with different methods. The +FM indication means that it is trained with NLL and FME; otherwise, it is trained with NLL only. The statistical values are the mean and twice the standard error for ten replication experiments.}
\label{table:metrics}
\end{table}

Table \ref{table:metrics} shows NLLs, force matching errors (FMEs), and reverse Kullback-Leibler divergences (KLDs) on the test set after ten training epochs for five different scenarios, with standard error for ten replication experiments. RQ-spline flow yielded the worst FME and KLD. It seems obvious that RQ-spline yielded the largest FME because $\mathcal{L}_{\text{FM}}$ can not be %is not 
used. %, but it is not obvious that 
RQ-spline also yielded the largest KLD, which %. This result 
suggests that if the normalized flow model itself is not $\mathcal{C}^2$-diffeomorphism, it may not be appropriate for Boltzmann generator even if $\mathcal{L}_{\text{FM}}$ is not used. %Designed to solve this problem, 
Smooth normalizing flow achieves great performance %metrics
regardless of using $\mathcal{L}_{\text{FM}}$.
Unlike RQ-spline flow, non-uniform B-spline flow achieved performance %metrics 
similar to smooth normalizing flow. One difference is that non-uniform B-spline flow yielded larger standard errors than smooth normalizing flow for some metrics. This is because non-uniform B-spline flow reaches similar metrics to smooth normalizing flow in most trials (about 9/10 trials) but has an outlier (about 1/10 trials). We discuss this phenomenon in the following section.

\begin{figure*}[ht!]
    \centering
    \includegraphics[width=\textwidth]{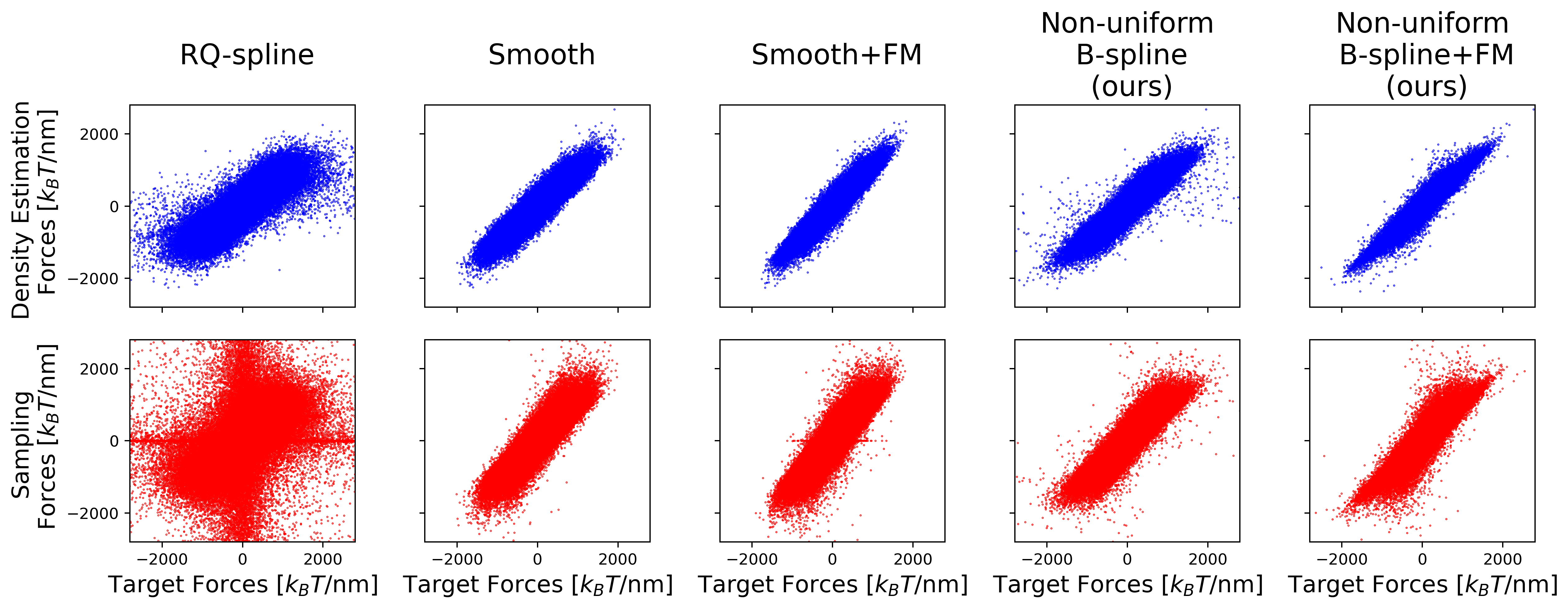}
    \caption{Scatterplot (10,000 samples) of forces estimated by each normalizing flow model (RQ-spline, smooth, and our non-uniform B-spline). The upper row is obtained by density estimation for test set samples, and the lower row is obtained by sampling with a flow model. The +FM indication means that it is trained with NLL + FM; otherwise, it is trained with NLL.} % Each scatterplot has .}
    \label{fig:forces}
\end{figure*}

Figure \ref{fig:forces} depicts the scatterplot of the forces estimated by the flow model in five different scenarios. The top row shows the results of density estimation, and the bottom row shows the results of sampling. There was no significant difference between the two rows except for RQ-spline flows, which showed the worst performance. The other four scenarios showed similar performance, but each achieved slightly better performance when using FM. Therefore, both smooth normalizing flow and non-uniform spline flow seem to model the force well.

\subsection{Dynamics Simulation by Density Estimation}
In the previous experiment, our non-uniform B-spline flow seemed to model force well. However, simulating molecular dynamics through density estimation using normalizing flows is much more challenging. Since the flow model does not have any internal algorithm that can suppress numerical errors, it quickly breaks the molecular structure unless it is very well-trained. Therefore, here we further verify our non-uniform B-spline flow thoroughly through the molecular dynamics simulation.

We used three models for molecular dynamics simulation; RQ-spline flow, smooth normalizing flow with FM, and non-uniform B-spline flow with FM. Each model is identical to the one used in Figure \ref{fig:forces}. 

\begin{figure}[t!]%
\centering
\subfloat[RQ-spline]{{\includegraphics[ width=0.48\columnwidth ]{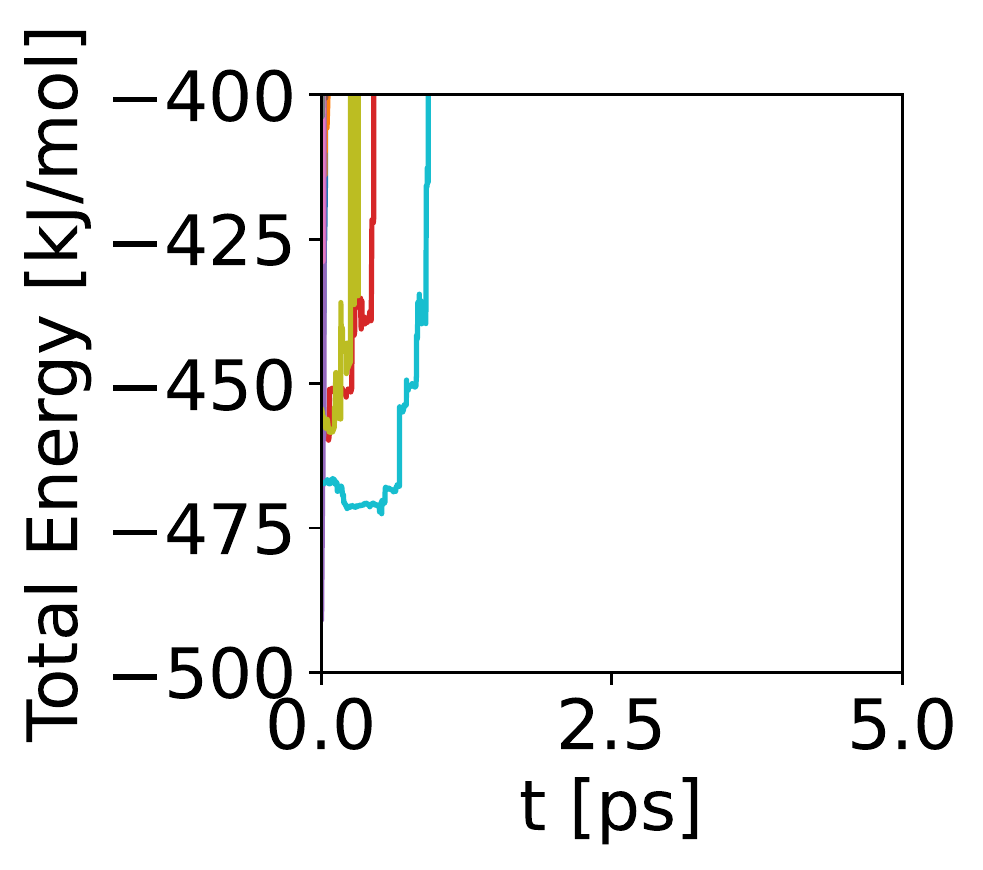} }}% 
\hfil
\subfloat[RQ-spline (rescaled)]{{\includegraphics[ width=0.40\columnwidth ]{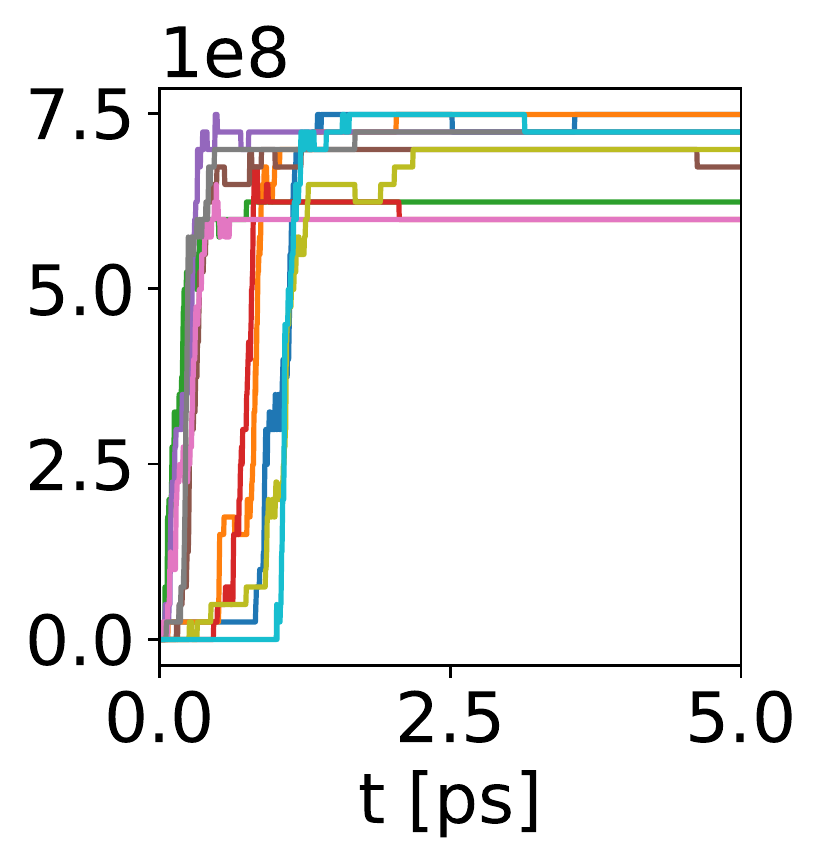} }}% 
\hfill
\subfloat[Non-uniform B-spline (ours)]{\makebox[0.48\columnwidth][r]{\includegraphics[width=0.44\columnwidth ]{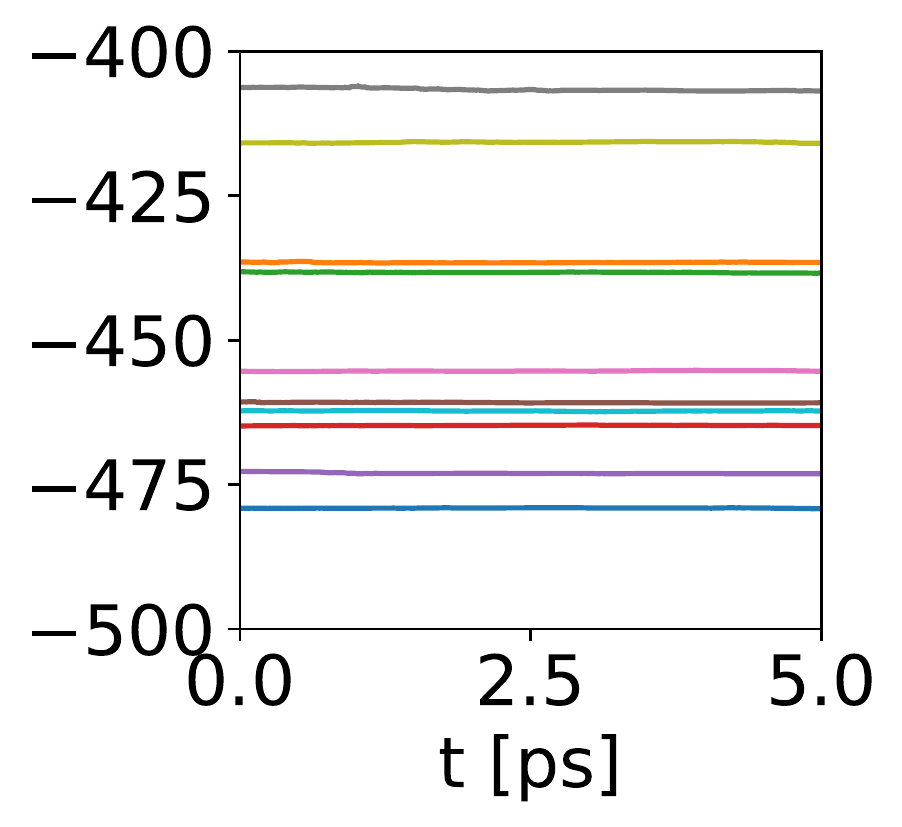} } }
\hfil
\subfloat[Smooth]{{\includegraphics[width=0.44\columnwidth ]{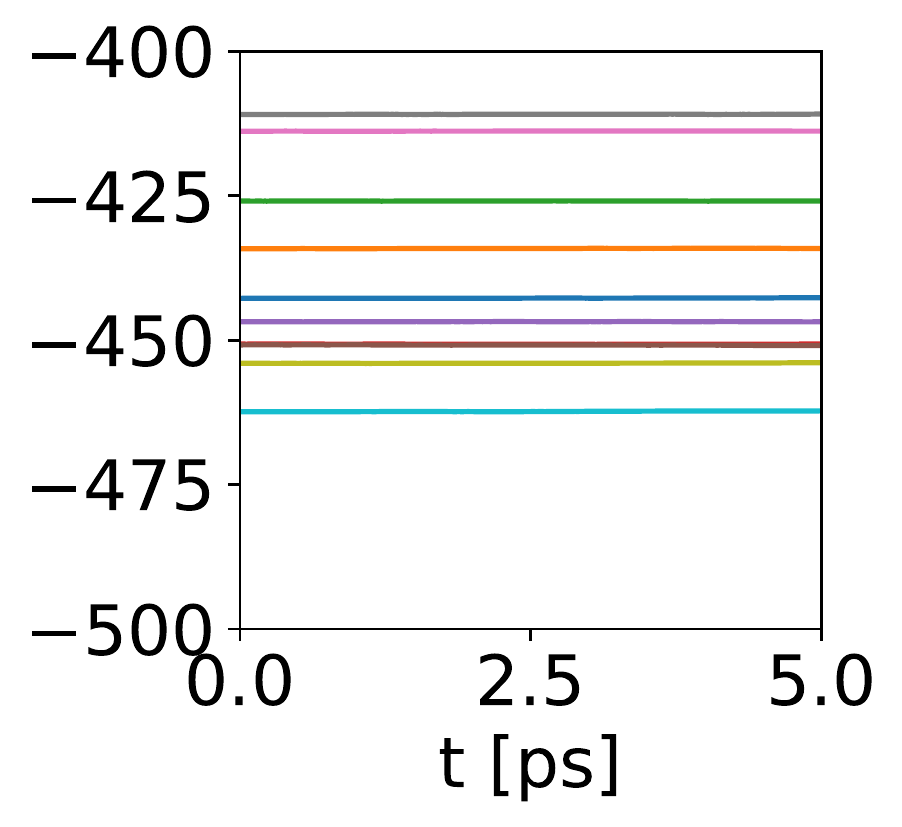} }}% 
\caption{Potential energies during the molecular dynamics simulation, estimated by (a) RQ-spline flow ((b) rescaled), (c) non-uniform B-spline flow, and (d) smooth normalizing flow, respectively. The simulation started with ten stable random initial configurations.}%
\label{fig:simulation}% 
\end{figure}

Figure \ref{fig:simulation} shows the potential energies during the molecular dynamics simulation. 
%Note that the y-axis scale in Figure \ref{fig:simulation} (b) is far larger than the others.
As in Figure 4 (a) and (b), RQ-spline flow failed to maintain the initial state for all ten different initial values. On the other hand, in Figure 4 (c) and (d), both non-uniform B-spline flow and smooth normalizing flow well-maintained all initial values.

\subsection{Runtime Comparison}
\begin{table}[t]
\small
\centering
\begin{tabular}{|l|c|c|c|}
\hline
Architecture          & \#params &\multicolumn{1}{l|}{\begin{tabular}[c]{@{}l@{}}Runtime \\ (Reverse)\end{tabular}} & \multicolumn{1}{l|}{\begin{tabular}[c]{@{}l@{}}Runtime\\ (Forward)\end{tabular}} \\ \hline
RQ-spline       & 285,497 & \begin{tabular}[c]{@{}c@{}}0.59\\ ($\pm$ 0.01)\end{tabular}                                             & \begin{tabular}[c]{@{}c@{}}0.49\\ ($\pm$ 0.01)\end{tabular}                                                                                          \\ \hline
Smooth          & 314,456 & \begin{tabular}[c]{@{}c@{}}19.8\\ ($\pm$ 0.26)\end{tabular}                 & \begin{tabular}[c]{@{}c@{}}0.64\\ ($\pm$ 0.01)\end{tabular}                         \\ \hline
\begin{tabular}[c]{@{}l@{}}Non-uniform \\ B-spline (ours)\end{tabular} & 380,982 & \begin{tabular}[c]{@{}c@{}}1.12\\ ($\pm$ 0.02)\end{tabular}                                                                                   & \begin{tabular}[c]{@{}c@{}}0.72\\ ($\pm$ 0.01)\end{tabular}                                                                                          \\ \hline
\end{tabular}
\caption{Runtimes per sample (in ms) of RQ-spline flows, smooth flows and non-uniform B-spline flows. The runtime is averaged over 10,000 samples each. The statistical values are the mean and twice the standard error for ten replication experiments. All computations were conducted on NVIDIA GeForce RTX3090.}
\end{table}
We compared the runtime of forward and reverse operations for the models used in the previous experiment. The forward operation performs density estimation, and the reverse operation performs sampling. Table 2 shows the average runtime for 10,000 samples. The table also shows the model's total number of parameters, indicating no significant difference in the size of the flow models. Because all three models performed forward operations analytically, they had similar runtimes (proportional to the number of parameters). % (We think if the models had more similar sizes, they would have had more similar speeds).

In reverse operation, since only smooth normalizing flow was non-analytic (black-box root-finding), smooth normalizing flow was the slowest. Non-uniform B-spline flow, whose reverse operates analytically, was about 17 times faster than smooth normalizing flow. This demonstrates that the proposed non-uniform B-spline flow has a great advantage in runtime over smooth normalizing flow. We also observed that RQ-spline flow had an increased runtime in reverse operation since RQ-spline flow and non-uniform B-spline flow solve the quadratic and cubic equations, respectively.

\section{Discussion}
\paragraph{Outlier issue in non-uniform cubic B-spline flows}
In the Boltzmann generator experiment, our non-uniform cubic B-spline flow model showed an outlier with a large reverse KLD, about once in ten times. One possible explanation for %We think that 
this phenomenon can be %comes from 
the numerical instability of the cubic equation root-finding formula \cite{durkan2019cubic}.
Since the implementation of the model uses a floating point operation, it is impossible to guarantee whether the root-finding formula will actually find the root correctly beyond machine precision. This problem has also been raised in cubic spline flows \cite{durkan2019cubic}, a study using the same root-finding formula as ours. Considering 32-bit floating-point precision, cubic spline flows clipped the input of every cubic-spline transformation to $[10^{-6}, 1-10^{-6}]$ instead of $[0,1]$. Similar trick of clipping the inputs or the increasement of 
%If we also use that kind of constraint or increase 
the floating-point precision (\textit{e.g.}, 64-bit or 128-bit) %, we 
may suppress outliers from the instability of the equation root formula for much more accurate physics system modeling.
% bi-directional training is worse in SNF ?

\paragraph{Limitations}
Spline-based normalizing flows, including our non-uniform B-spline flows, are much more computationally burdensome than affine coupling flows. For this reason, normalization flows dealing with images mainly employ affine transformations. Researchers in various fields such as medical imaging have studied invertibillity \cite{chun2009simple,sdika2013sharp} and fast transformation \cite{unser1993b1, unser1993b2} of uniform B-splines. Using uniform B-splines rather than non-uniform B-splines, it may be possible to construct normalizing flows as fast as affine coupling flows.

\section{Conclusion}
%In this paper, 
We proposed non-uniform B-spline flows based on the sufficient conditions that  %. We demonstrated that 
$k$th-order non-uniform B-spline transformation is a $\mathcal{C}^{k-2}$-diffeomorphism, by proving bi-Lipschitz continuity and surjectivity on various compact domains. Experiments demonstrated that our non-uniform B-spline flows can solve the force matching problem in Boltzmann generators better than previous spline-based flows and as good as smooth flows. Our method can admit %'s capability of admitting 
analytic inverses so that it is much faster than smooth  flows.

\section{Acknowledgements}

This work was supported by the National Research Foundation of Korea(NRF) grants funded by the Korea government(MSIT) (NRF-2022R1A4A1030579, NRF-2022M3C1A309202211) and by Basic Science Research Program through the National Research Foundation of Korea(NRF) funded by the Ministry of Education(NRF-2017R1D1A1B05035810). Also, the authors acknowledged the ﬁnancial supports from BK21 FOUR program of the Education and Research Program for Future ICT Pioneers, Seoul National University.

%\section{Supplementary Material :\\ Non-uniform B-spline Flows}

\appendix \section{Appendices} 
\section{Experimental Details}
\subsection{Illustrative toy example}
We provide additional informations about the illustrative toy example experiment.
\subsubsection{Results for modelling periodic probability density}
\begin{figure}[h]%
\centering
\begin{tabular}{ccc}
\begin{tabular}[c]{@{}c@{}}(a) Ground\\ Truth\end{tabular}                                                   & \multicolumn{2}{c}{\adjustbox{valign=m}{{\includegraphics[width=0.7\columnwidth ]{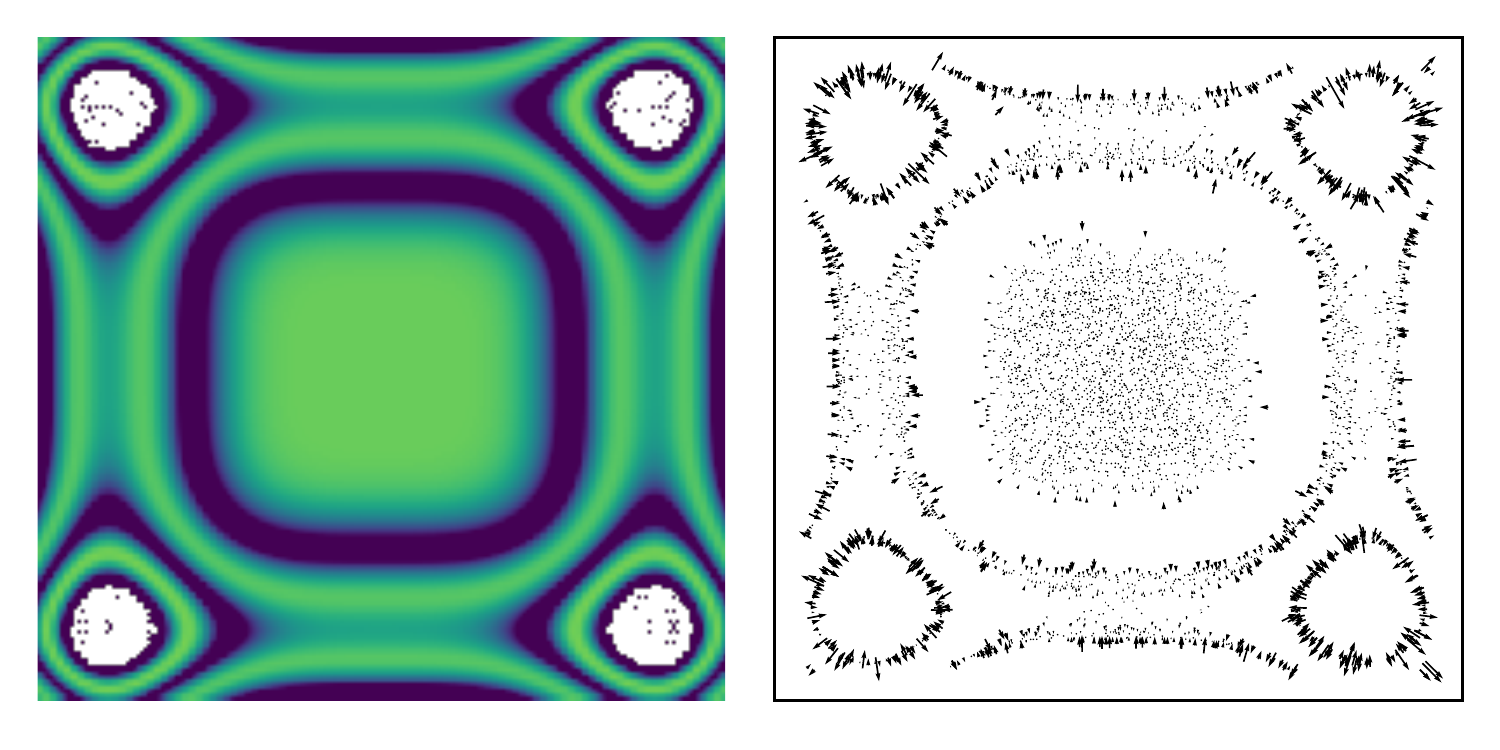} }}} \\
\begin{tabular}[c]{@{}c@{}}(b) RQ-\\spline\end{tabular}                                                      & \multicolumn{2}{c}{\adjustbox{valign=m}{{\includegraphics[width=0.7\columnwidth ]{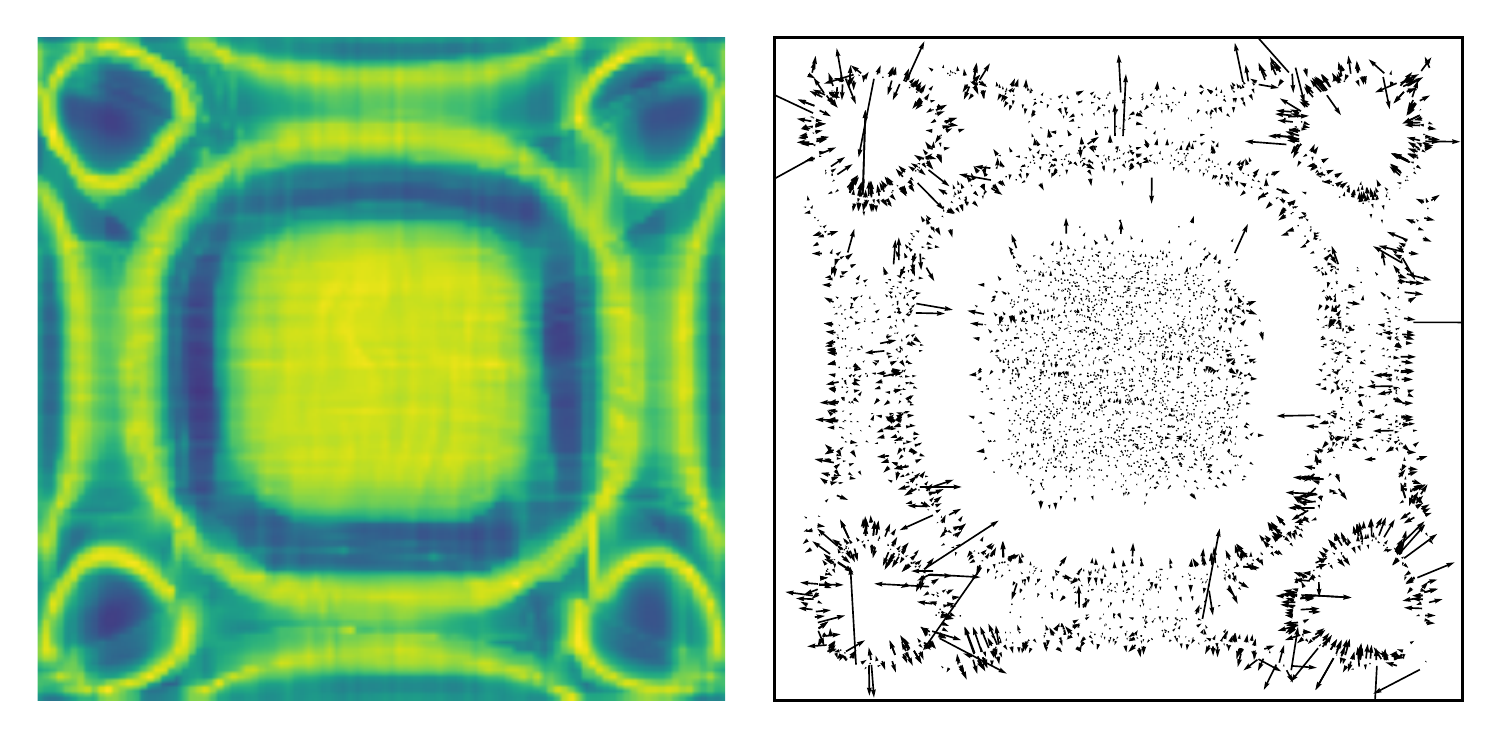} }}} \\
(c) Smooth                                                         & \multicolumn{2}{c}{\adjustbox{valign=m}{{\includegraphics[width=0.7\columnwidth ]{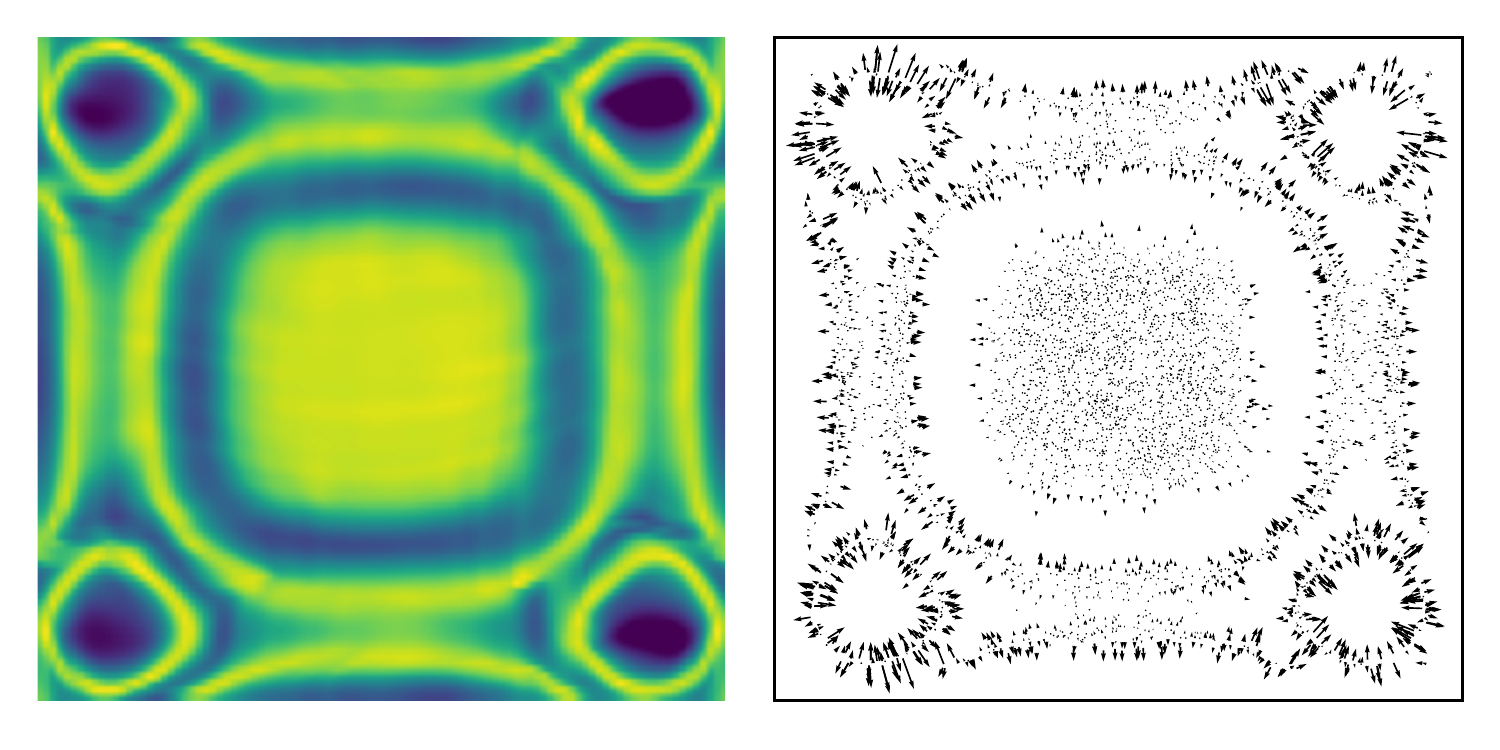} }}} \\
\begin{tabular}[c]{@{}c@{}}(d) Non-\\uniform\\ B-spline \\(ours)\end{tabular} & \multicolumn{2}{c}{\adjustbox{valign=m}{{\includegraphics[width=0.7\columnwidth ]{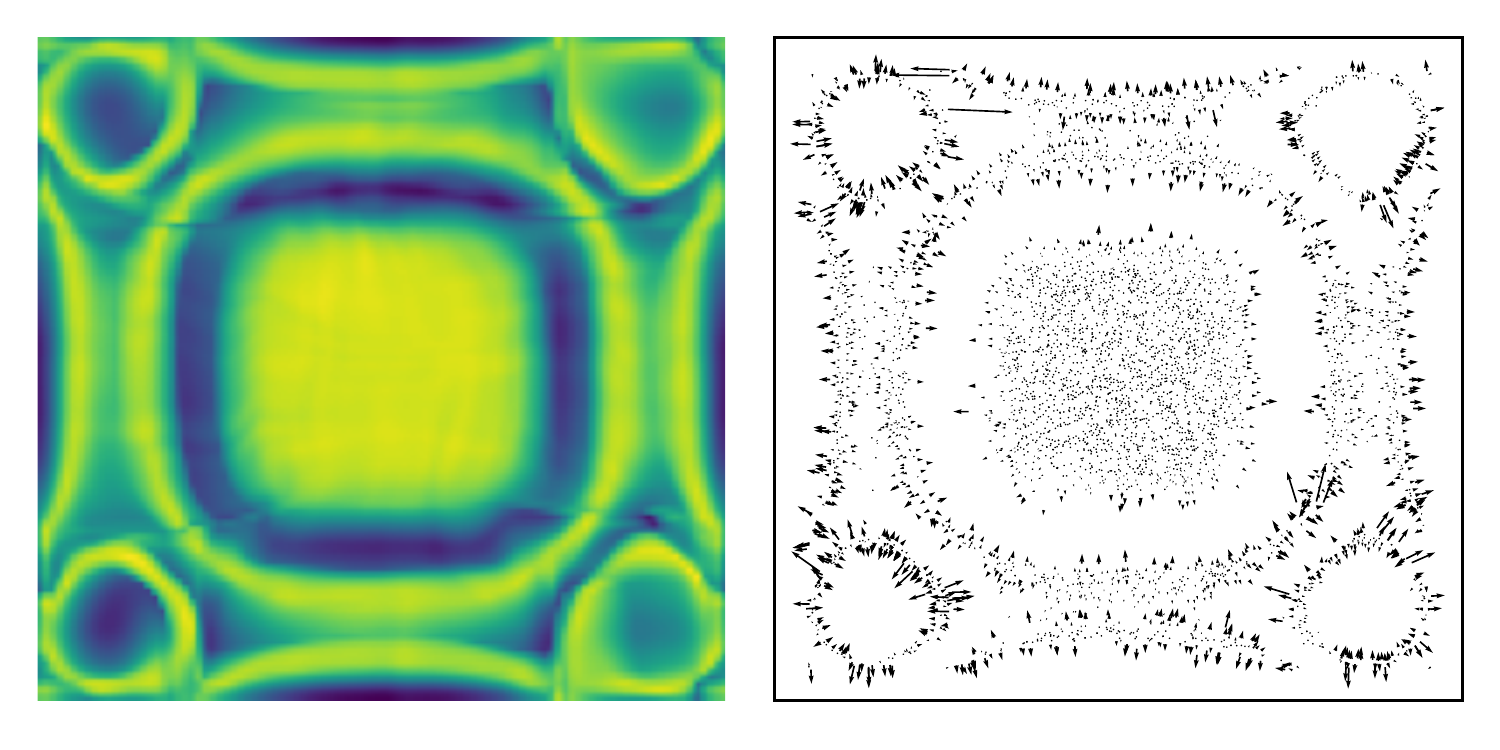} }}}\\
\end{tabular}
\caption{Probability density (left) and corresponding force field (right) of (a) ground truth, approximated by (b) RQ-spline flows, (c) Smooth flows, and (d) Non-uniform (cubic) B-spline flows (ours). }%
\label{fig:toy_periodic}% 
\end{figure}

We also provide 2D Toy experimental results modeling periodic probability density in Fig. \ref{fig:toy_periodic}. 

\subsubsection{Probability density functions and dataset}
In this experiment, we used simple ground truth probability densities previously used in smooth normalizing flows \cite{kohler2021smooth}. 
For the aperiodic case as in Fig. \ref{fig:toy}, we used the following probability density
\begin{equation}
    p_\mathbf{x}^\star (\mathbf{x}) \propto A_i \exp \left(-\frac{\lVert \lVert \mathbf{x} \rVert_2 - R_i \lVert_2^2}{2\sigma}\right)
\end{equation}
with $A = [1,0.8,0.6,0.4]$ and $\sigma=0.06$.
For the periodic case as in Fig. \ref{fig:toy}, we used the following probability density
\begin{equation}
\begin{aligned}
    C(\mathbf{x}) & = \left( \cos (x_1^2) + \cos (x_2^2) +2 \right)^\frac{1}{2}\\
    p_\mathbf{x}^\star (\mathbf{x}) & \propto A_i \exp \left(-\frac{\lVert C(\mathbf{x}) - R_i \lVert_2^2}{2\sigma}\right)     
\end{aligned}
\end{equation}
with $A = [1,0.8,0.6,0.4]$ and $\sigma=0.05$.

To make a dataset $\mathcal{X}$, we firstly ran Metropolis-Hastings for 1,000 steps, using 10,000 parallel chains. We then operated each chain for ten more steps to provide a total of 100,000 samples.

\subsubsection{Architecture}
We employ the same structure as in smooth normalizing flows except for some hyperparameters. We used four coupling layers, and we swapped the first and second dimensions between each coupling layer. Inside the coupling layers, B-spline and RQ-spline transformations had 32 bins (i.e., $s-r = 32$), and smooth normalizing flow had 32 mixture components. NN in coupling layers was a fully connected network with two hidden layers with a width of 100. When the probability density is periodic, we applied cosine to the input of NN to make the model periodic. For our non-uniform B-spline flow, We set $\epsilon_t = \epsilon_\alpha = 10^{-4}$.

\subsubsection{Training}
We used $\mathcal{L}_{\text{NLL}}$ as a loss function. We trained the network using Adam~\cite{kingma2014adam} optimizer (with learning rate = 0.0005, batch size = 1,000), for 8,000 epochs. 

\subsection{Boltzmann generator}
\subsubsection{Dataset}
In this experiment, the Boltzmann generator models one alanine dipeptide molecule contained in the implant solvent. The ground truth energy was obtained using the Amber ff99SB-ILDN force field with the Amber99 Generalized Born (OBC) solvation model, as same in smooth normalizing flows~\cite{kohler2021smooth}. Covalent bonds were assumed to be flexible. The training set (i.e., $\mathcal{X}$) was generated by molecular dynamics simulations in OpenMM. In the simulation, the time step was 1 fs; simulation time was 1 $\mu$s. Langevin integration was performed with a 1/ps friction coefficient, and data were collected at 1-ps intervals.

\subsubsection{Architecture}
In our experiment, the Boltzmann gernerator models the structure of alanine dipeptide, which is a 22-atom molecule. Torsions, angles and bond lengths represent the structure of alanine dipeptide. They have degrees of freedom of 19, 20, and 21, respectively. Since angles and bonds should be physically reasonable, we assumed that angles were constrained in [0.15$\pi$, $\pi$] and bond lengths were in [0.05 nm, 0.3 nm]. Under this assumption, torsion, angle, and bond were isomorphic to $\mathcal{S}^1, \mathbb{I}$, and $\mathbb{I}$, respectively. Therefore, we set $\Omega = \mathbb{T}^{19} \times \mathbb{I}^{20} \times \mathbb{I}^{21}$ and $\mathbf{u} \sim \mathrm{Uniform}(\Omega)$.

\begin{figure*}[t]
    \centering
    \includegraphics[width=0.80\textwidth]{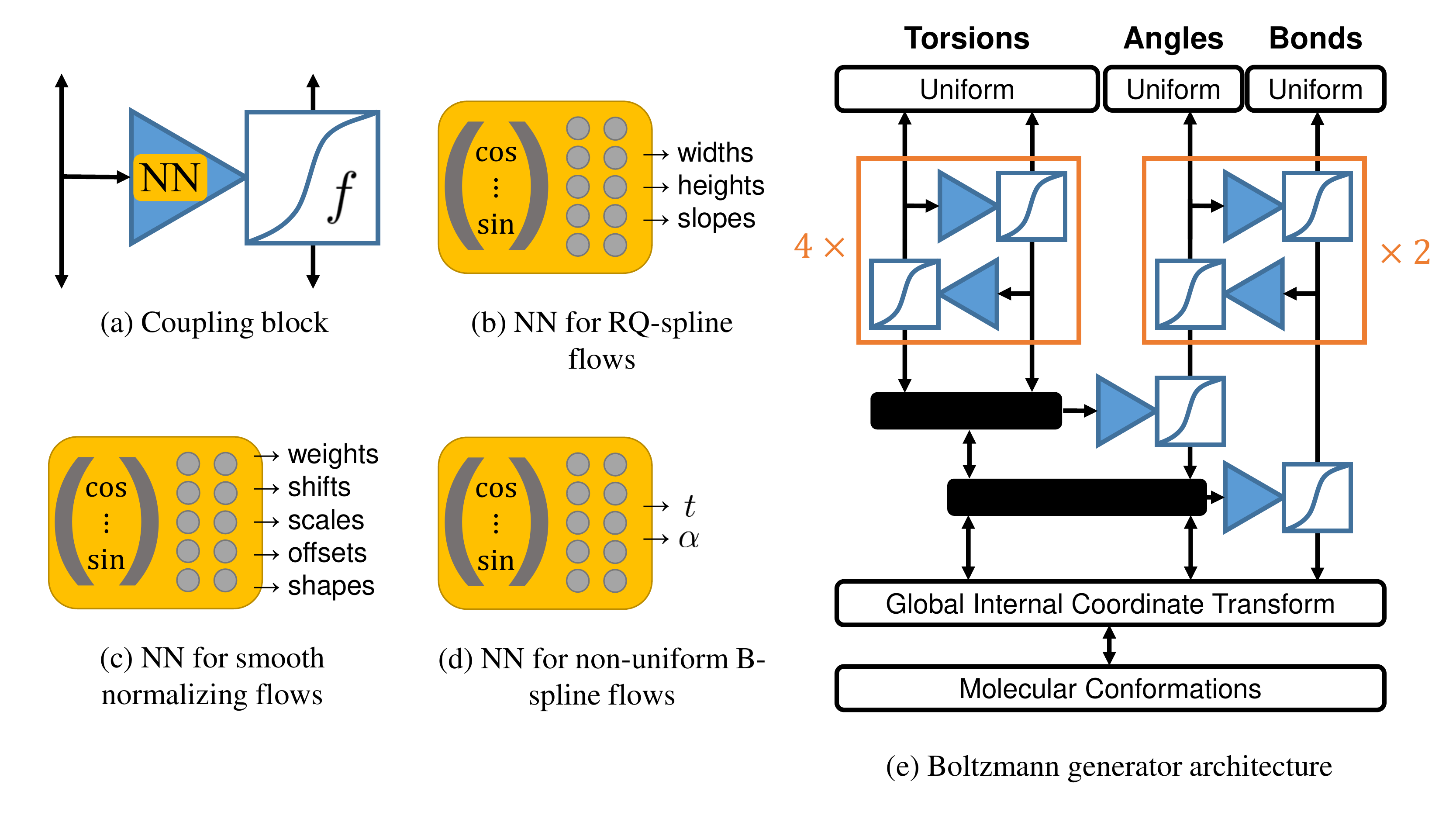}
    \caption{Network architecture - (a) Coupling block : using a part of the input, a fully connected network NN generates parameters of an elementwise diffeomorphic transformation f. Then f is applied to the other part of the input. (b) NN for RQ-spline transformations~\cite{durkan2019neural} generates widths, heights, and slopes. (c) NN for smooth normalizing flows~\cite{kohler2021smooth} generates weights, shifts, scales, offsets, and shapes. (d) NN for non-uniform B-spline flows (ours) generates $t$ and $\alpha$. (e) Boltzmann generator architecture. Torsions are periodic (i.e., $\mathcal{S}^1$), angles and bonds are aperiodic (i.e., $\mathbb{I}$). }
    \label{fig:architecture}
\end{figure*}

We constructed a network with coupling blocks in Figs. \ref{fig:architecture} (a). As shown in Equation \eqref{eq_flow}, NN generates parameters of $f$. Generated parameters are different depending on the type of $f$, as shown in Figs. \ref{fig:architecture} (b), (c), and (d).
For a fair runtime comparison, we designed the networks so that each algorithm's total number of parameters is similar (see Table 2).
To this end, for RQ-spline transformations, the number of bins was 16. For mixture of bump functions in smooth normalizing flows, the number of mixture components was 8. For our non-uniform B-spline transformations, the number of bins was 32. NNs had two hidden fully connected layers with widths of 64 and $\sin$ activations. For our non-uniform B-spline flow, We set $\epsilon_t = \epsilon_\alpha = 10^{-6}$.

Figure \ref{fig:architecture} (e) shows the structure of the entire network. Since torsions were defined in $\mathbb{T}^{19}$, we employed eight periodic coupling layers. Angles and bonds were defined in $\mathbb{I}^{20}$ and $\mathbb{I}^{21}$, respectively, so four aperiodic coupling layers were employed. After that, an aperiodic coupling transformation was performed on angles using the merged torsions as conditional variables. Finally, an aperiodic coupling transformation was performed on bonds using the merged torsions and angles as conditional variables.

\subsubsection{Training}
We trained the network using Adam~\cite{kingma2014adam} optimizer (with learning rate = 0.0005, batch size = 128), for 10 epochs. We scaled the learning rate by 0.7 every epoch. The training set was 90\% of the dataset, and we computed the metrics over the rest of the dataset.

\subsubsection{Additional results}
\begin{figure*}[t!]
    \centering
    \includegraphics[width=0.88\textwidth]{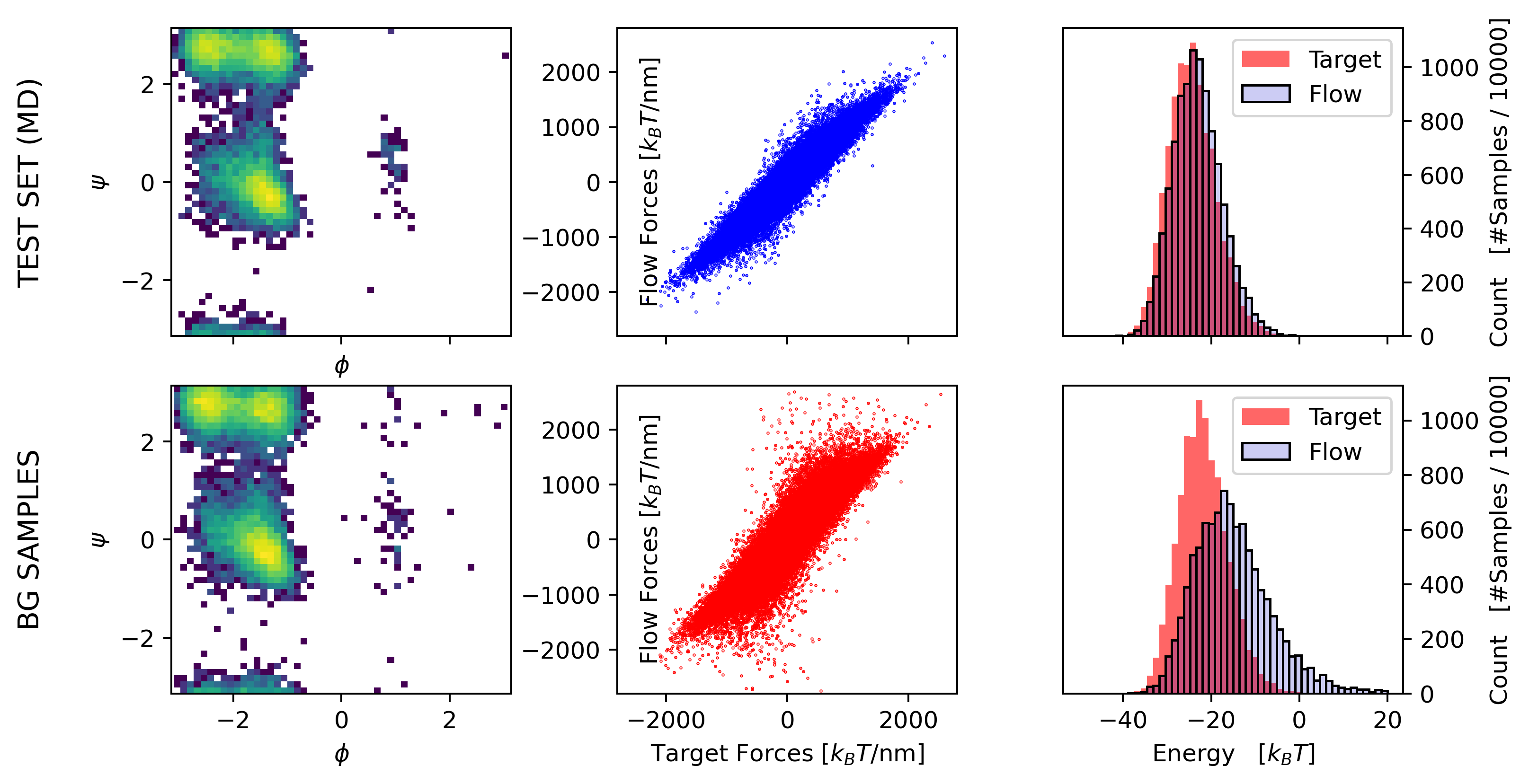}
    \caption{Neural diffeomorphic non-uniform B-spline flow (ours) trained on alanine dipeptides with negative log-likelihood and force matching error as a loss function. The top and bottom rows show performance for 10,000 samples from the test set and flow models, respectively. The left column shows the joint distributions of backbone torsion angles. The middle column shows scatter plots of the force calculated by the flow model vs. the ground truth. The right column shows energy histograms for the energy calculated by the flow model and the ground truth. (Note that the flow energy was shifted to match its minimum value with the target energy.)}
    \label{fig:bspline_fm0_bg}
\end{figure*}

\begin{figure*}[t!]
    \centering
    \includegraphics[width=0.88\textwidth]{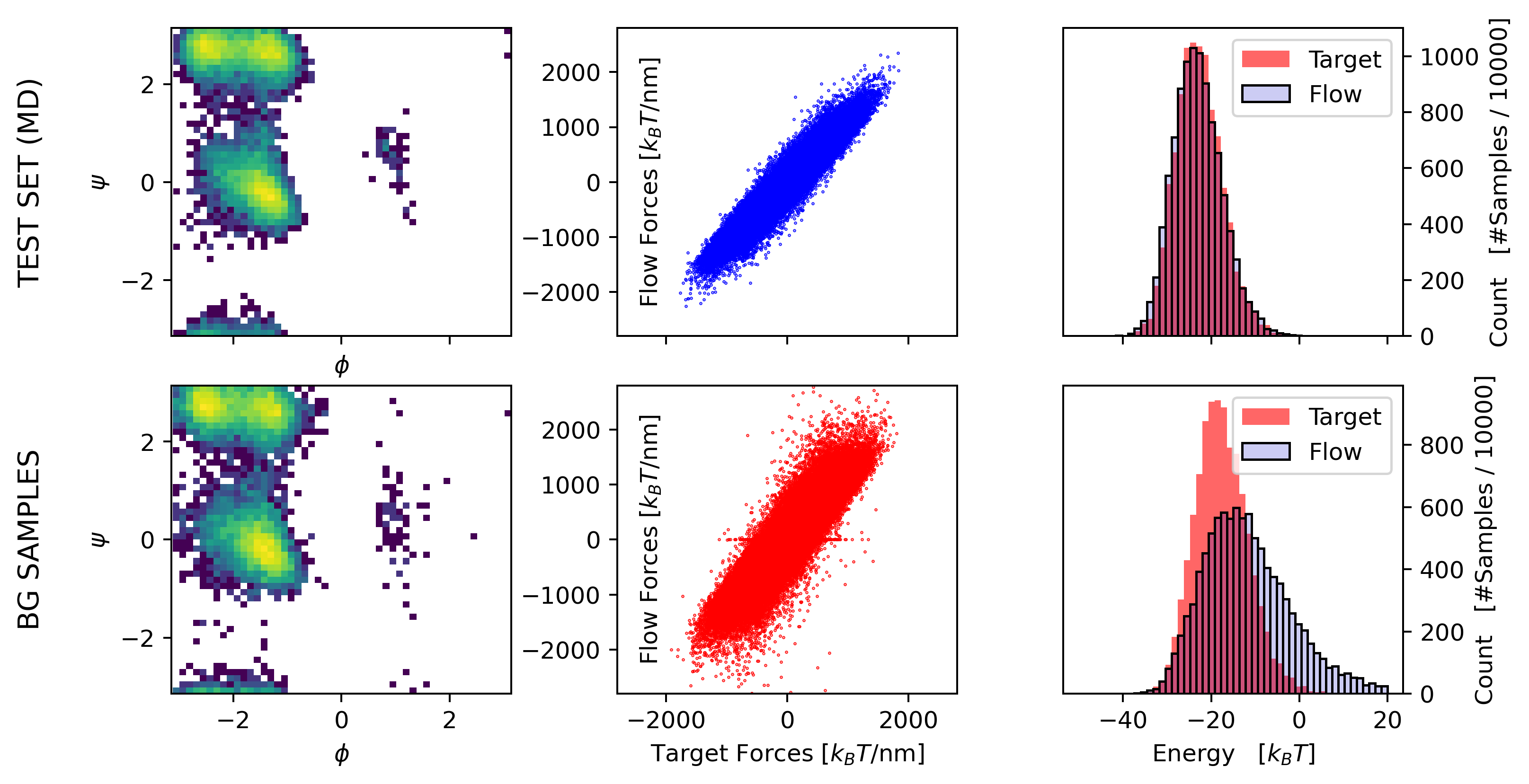}
    \caption{Smooth normalizing flow~\cite{kohler2021smooth} trained on alanine dipeptides with negative log-likelihood and force matching error as a loss function. The top and bottom rows show performance for 10,000 samples from the test set and flow models, respectively. The left column shows the joint distributions of backbone torsion angles. The middle column shows scatter plots of the force calculated by the flow model vs. the ground truth. The right column shows energy histograms for the energy calculated by the flow model and the ground truth. (Note that the flow energy was shifted to match its minimum value with the target energy.)}
    \label{fig:smooth_fm0_bg}
\end{figure*}

\begin{figure*}[t!]
    \centering
    \includegraphics[width=0.88\textwidth]{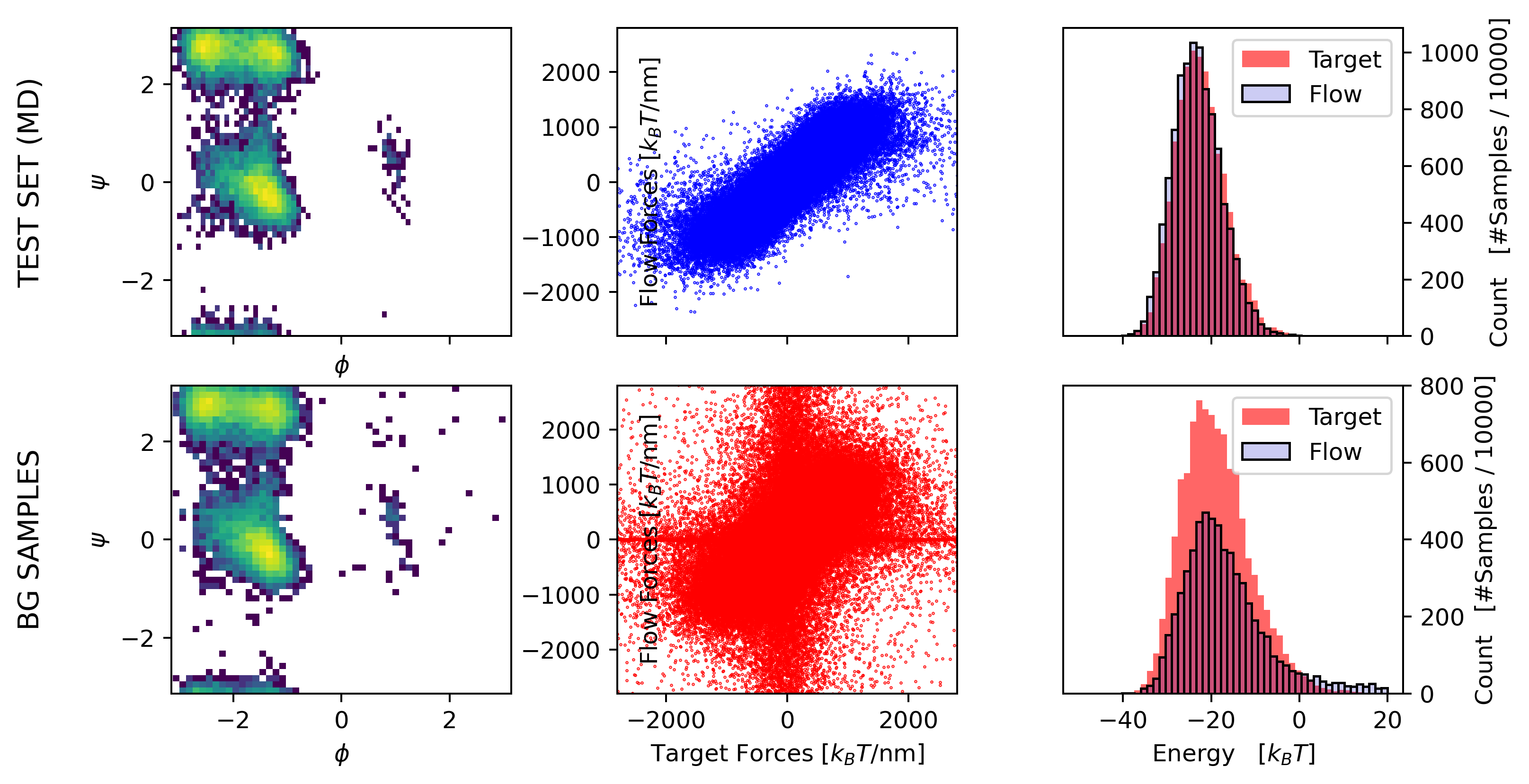}
    \caption{RQ-spline flow~\cite{durkan2019neural} trained on alanine dipeptides with negative log-likelihood and force matching error as a loss function. The top and bottom rows show performance for 10,000 samples from the test set and flow models, respectively. The left column shows the joint distributions of backbone torsion angles. The middle column shows scatter plots of the force calculated by the flow model vs. the ground truth. The right column shows energy histograms for the energy calculated by the flow model and the ground truth. (Note that the flow energy was shifted to match its minimum value with the target energy.)}
    \label{fig:rqspline0_bg}
\end{figure*}

Figures \ref{fig:bspline_fm0_bg},\ref{fig:smooth_fm0_bg},\ref{fig:rqspline0_bg} depicts additional results of the Boltzmann generator experiments using non-uniform B-spline flow (ours), smooth normalizing flow~\cite{kohler2021smooth}, and RQ-spline flow~\cite{durkan2019neural}, respectively. The top and bottom rows show performance for 10,000 samples from the test set and flow models. The left column shows the joint distributions of backbone torsion angles. The middle column shows scatter plots of the force calculated by the flow model vs. the ground truth. The right column shows energy histograms for the energy calculated by the flow model and the ground truth. We shifted the flow energy to match its minimum value with the target energy.


\subsubsection{Molecular dynamics simulation}
The simulation started with ten stable random initial configurations from the dataset and random velocities from Maxwell-Boltzmann distribution at 300 K. The simulation was equilibrated for the first 5 ps using a Langevin thermostat with 10/ps friction coefficient. Then, the thermostat was removed, which made the simulation may have difficulty suppressing the numerical errors. The simulation time step was 0.1 fs and the simulation was performed for 5 ps.

\section{Additional Resources}
\subsection{Computing requirements}
We used one Intel(R) Core(TM) i9-11900KF CPU and one NVIDIA GeForce GTX 3090 GPU for all experiments. We ran our experiments on Ubuntu 18.04 LTS. Training could be carried out in parallel (up to 4) in this environment. Since it took about 40 minutes per epoch, We could complete the training in about 400 minutes. Since we trained 5 models $\times$ 10 replicas, it took approximately 5$\times$10$\times$400$/$4 = 5000 minutes. 
\subsection{Third-party software}
We used the source code of smooth normalizing flows~\cite{kohler2021smooth} and neural spline flows~\cite{durkan2019neural}. 
We used OpenMM for the simulations, and mdtraj to analyze molecular dynamics. We list the third-party software we used as follows.

\begin{itemize}
    \item{OpenMM :} Eastman, P.; Swails, J.; Chodera, J. D.; McGibbon, R. T.;
Zhao, Y.; Beauchamp, K. A.; Wang, L.-P.; Simmonett, A. C.;
Harrigan, M. P.; Stern, C. D.; et al. 2017. OpenMM 7: Rapid
development of high performance algorithms for molecular
dynamics. PLoS computational biology, 13(7): e1005659.
    \item{mdtraj :} McGibbon, R. T.; Beauchamp, K. A.; Harrigan, M. P.; Klein,
C.; Swails, J. M.; Hern\'andez, C. X.; Schwantes, C. R.;
Wang, L.-P.; Lane, T. J.; and Pande, V. S. 2015. MDTraj:
a modern open library for the analysis of molecular dynam-
ics trajectories. Biophysical journal, 109(8): 1528–1532.
\end{itemize}

% Use \bibliography{yourbibfile} instead or the References section will not appear in your paper
\bibliography{refs}

\end{document}